\documentclass[10pt,twocolumn,letterpaper]{article}

\usepackage{cvpr}              

\usepackage[linesnumbered, boxed, ruled]{algorithm2e}
 \RestyleAlgo{ruled}
 \SetCommentSty{mycommfont}
 \usepackage{amsmath}
 \usepackage{multirow}
 \usepackage{pifont}
 \newcommand{\cmark}{\textcolor{green}{\ding{51}}}%
\newcommand{\xmark}{\textcolor{red}{\ding{55}}}%
\usepackage{commath}
\usepackage[dvipsnames]{xcolor}
\usepackage{soul}

\definecolor{my_yellow}{HTML}{FFC500}

\newcommand{\te}{\text{TransErr}}

\definecolor{cvprblue}{rgb}{0.21,0.49,0.74}
\usepackage[pagebackref,breaklinks,colorlinks,allcolors=cvprblue]{hyperref}

\def\paperID{4955} 
\def\confName{CVPR}
\def\confYear{2025}

\title{Fast Multi-view Consistent 3D Editing with Video Priors}

\author{Liyi Chen \quad Ruihuang Li \quad Guowen Zhang \quad Pengfei Wang \quad Lei Zhang\thanks{Corresponding author.} \\
The Hong Kong Polytechnic University \\
{\tt\small \{liyi0308.chen, guowen.zhang, pengfei.wang\}@connect.polyu.hk }\\ 
{\tt\small \{csrhli,cslzhang\}@comp.polyu.edu.hk}\\
Project Page: \url{https://mt-cly.github.io/ViP3DE}
}

\begin{document}
\maketitle
\begin{abstract}
Text-driven 3D editing enables user-friendly 3D object or scene editing with text instructions. Due to the lack of multi-view consistency priors, existing methods typically resort to employing 2D generation or editing models to process each view individually, followed by iterative 2D-3D-2D updating. However, these methods are not only time-consuming but also prone to over-smoothed results because the different editing signals gathered from different views are averaged during the iterative process.
In this paper, we propose generative \textbf{Vi}deo \textbf{P}rior based \textbf{3D E}diting (\textbf{ViP3DE}) to employ the temporal consistency priors from pre-trained video generation models for multi-view consistent 3D editing in a single forward pass. 
Our key insight is to condition the video generation model on a single edited view to generate other consistent edited views for 3D updating directly, thereby bypassing the iterative editing paradigm. Since 3D updating requires edited views to be paired with specific camera poses, we propose \textit{motion-preserved noise blending} for the video model to generate edited views at predefined camera poses.
In addition, we introduce \textit{geometry-aware denoising} to further enhance multi-view consistency by integrating 3D geometric priors into video models.
Extensive experiments demonstrate that our proposed ViP3DE can achieve high-quality 3D editing results even within a single forward pass, significantly outperforming existing methods in both editing quality and speed.

\end{abstract}    

\section{Introduction}
\label{sec:intro}

\begin{figure}[t]

    \centering
    \begin{minipage}[c]{1\linewidth}
    \includegraphics[width=1\linewidth]{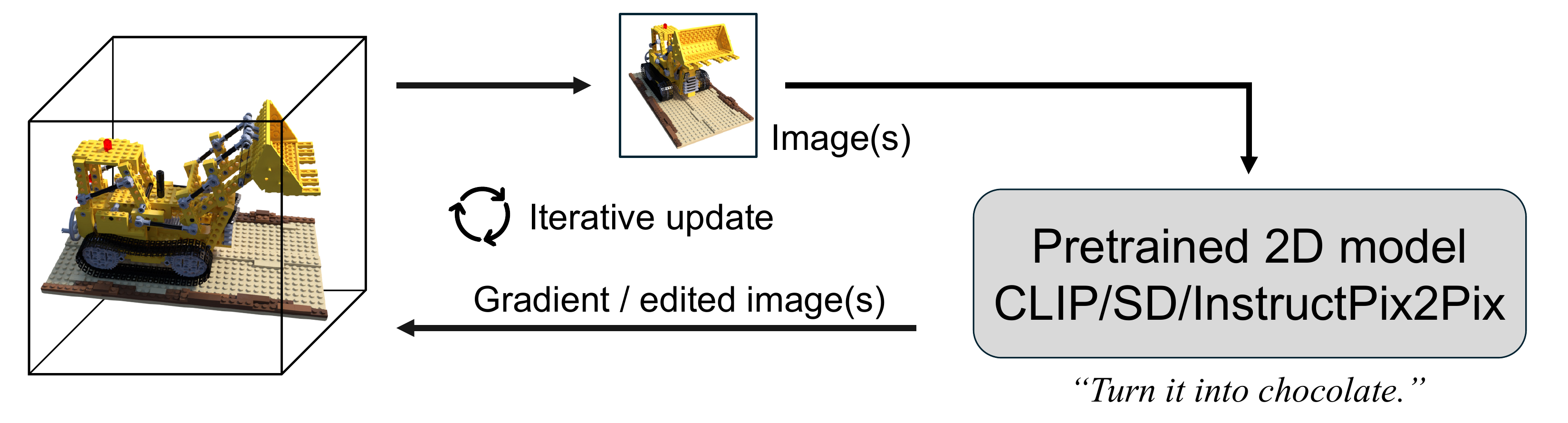}
    \end{minipage}
        \centering
    (a) \small{2D image prior based 3D editing.}
    \vspace{5pt}
    
    \centering
    \begin{minipage}[c]{1\linewidth}
    \includegraphics[width=1\linewidth]{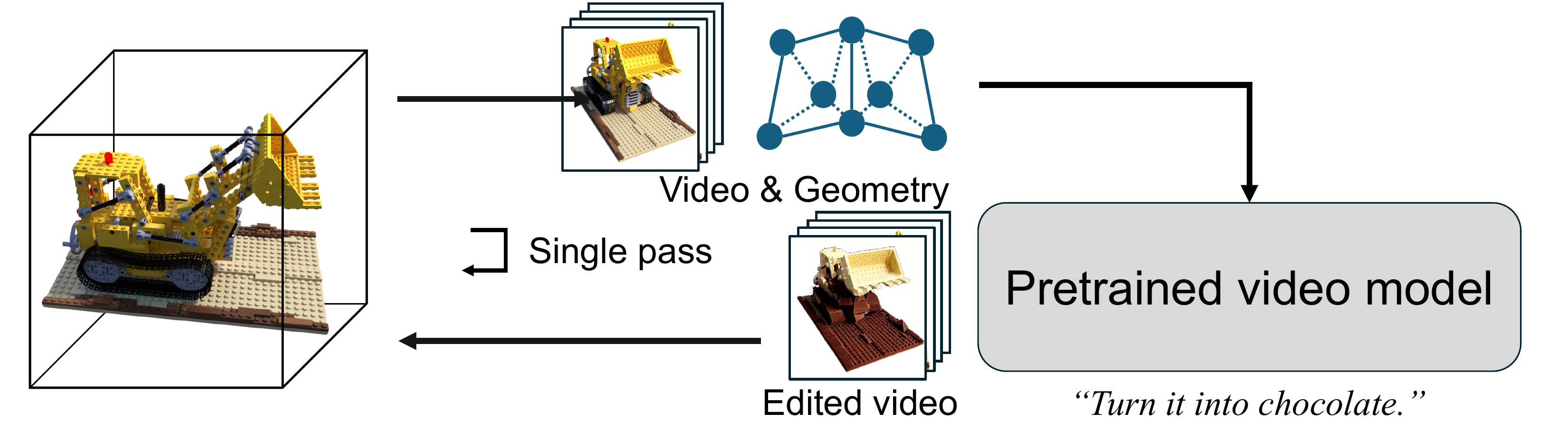}
    \end{minipage}
    \vspace{5pt}
    
    \centering
    (b) \small{ViP3DE integrates video priors and 3D priors for 3D editing.}

    \caption{\textbf{Motivation of ViP3DE.} (a) Most existing studies~\cite{instructn2n, shap-editor, clip-nerf, genn2n, dge, syncnoise, consistdreamer} employ pre-trained 2D models to iteratively update 3D assets, suffering from slow convergence and over-smoothed textures. (b) ViP3DE integrates video priors and source 3D priors to achieve multi-view consistent editing with a single pass.} 
\label{fig:teaser}
\end{figure}

3D editing aims to achieve high-quality personalized editing of 3D objects or scenes by modifying their shape and content.
Traditional 3D editing requires professional skills and tools to manipulate 3D mesh~\cite{editmesh1,editmesh2} or point clouds~\cite{editpoint1}, which is time-consuming and labor-intensive.
Recent development of 3D representations and multi-modality 2D models has revolutionized 3D editing. 
By integrating NeRF~\cite{nerf} or 3D Gaussians Splatting (GS)~\cite{3dgs} with off-the-shelf 2D multi-modal models such as CLIP~\cite{clip} and Stable Diffusion (SD)~\cite{sd}, a user-friendly 3D editing interface can be built using instructions.

Most existing methods employ 2D models to perform 3D editing with an iterative updating paradigm, as shown in Fig.~\ref{fig:teaser}(a). In each iteration, images from a randomly selected camera pose are rendered and edited individually using models like CLIP and SD~\cite{instruct3d23d,clip-nerf,nerfart,shap-editor,genn2n,focaldreamer} via Score Distillation, $e.g.,$ SDS, DDS, and SSD ~\cite{dds, ssd, editp23}, or 2D editor to output edited views~\cite{instructn2n,gaussianeditor}. The source 3D representation is updated with edited views through differential volume rendering~\cite{nerf} or rasterization~\cite{3dgs}.
Due to the lack of multi-view consistency priors, these per-image editing-based methods usually demand hundreds or thousands of iterations to average out the inconsistent gradient signals or the edited pixel values, resulting in slow convergence and over-smoothed texture. Although recent efforts have been proposed to synchronize different views by introducing extrapolated cross-attention~\cite{dge, consistdreamer} or point correspondence~\cite{syncnoise,efficientn2n}, they fail to achieve editing in a single pass, and these issues remain.

Motivated by the capability of pre-trained video models to generate inter-frame continuous videos, we propose to leverage pre-trained video priors to achieve multi-view consistent 3D editing in a single pass, as shown in Fig.~\ref{fig:teaser}(b). 
However, generative video models cannot be applied directly for 3D editing due to two issues. 
Firstly, 3D updating requires pairs of camera poses and edited images, while existing video models cannot produce edited images corresponding to precise camera poses~\cite{cameractrl,anyv2v}. Secondly, video models have limited understanding of 3D geometry and physics~\cite{sora, physic}. Therefore, the edited views often suffer from shape deformation or color shifts, which will cause undesired 3D editing results.  

In this paper, we propose \textbf{Vi}deo \textbf{P}rior based \textbf{3D E}diting (\textbf{ViP3DE}) to overcome these challenges. 
To obtain paired edited views and camera poses, we first render a source video along a known camera trajectory. We then acquire inverted video noise to guide subsequent edited view generation via an inversion-based paradigm. Note that, different from previous inversion-based video editing methods~\cite{motionclone, videoshop, anyv2v}, which either overestimates or underestimates the importance of the inverted noise and results in unsatisfactory camera motion or visual quality, we propose \textit{motion-preserved noise blending} to produce desired edited views under given 3D perspectives by blending inverted noise with random Gaussian noise as  initial noise.
Besides, to further improve the 3D consistency of edited views,  we propose \textit{geometry-aware denoising} to integrate 3D priors and video priors during diffusion process. We first build latent feature correspondence between the conditional view and other views based on the geometric relations between 3D representation and camera poses, which can then provide explicit constraints in video latent space across each denoising step. 
These two novel designs enable the video model to produce edited views that are continuous and consistent in 3D geometry. Finally, the source 3D asset is updated using the edited views in a single forward pass.

Our contributions are summarized as follows.
\begin{itemize}
    \item We propose ViP3DE, an early and pioneering work that introduces generative video priors for text-driven 3D editing in a training-free manner. 
     \item We introduce two novel designs, \textit{i.e.}, motion-preserved noise blending and geometry-aware denoising, to produce 3D-consistent edited views with high-quality visual results in a single pass.
    \item Extensive experiments demonstrate that ViP3DE significantly outperforms previous methods in both efficiency and editing quality.
\end{itemize}

\section{Related Work}
\label{sec:related_works}

\textbf{Text-driven 3D Editing}.
With the advancement of 2D multi-modal models~\cite{diff4splat,ir3dbench} such as CLIP~\cite{clip}, SD~\cite{sd}, and InstructPix2Pix~\cite{instructpix2pix}, text-driven 3D editing has attracted much attention recently. 
One line of research aims to translate source 3D representations to edited 3D representations directly under the guidance of off-the-shelf 2D models. 
CLIP-NeRF~\cite{clip-nerf} and NeRF-Art~\cite{nerfart} optimize the NeRF by employing CLIP as the discriminator. 
Shape-Editor~\cite{shap-editor} learns a NeRF latent mapping using text-to-image models with SDS loss~\cite{dreamfusion}.
GenN2N~\cite{genn2n} builds a translated NeRF that takes image embeddings as input and generates edited novel views in specific poses.
Other works along this line~\cite{instruct3d23d, progressive3d, focaldreamer,vox,dreameditor,proteusnerf,edit_diffnerf,related_1,related_2} share a similar idea by employing different variants of SDS with different 2D models. 
Due to the inherited limitation of SDS loss, however, these methods usually produce over-saturated and low-quality editing results.

Another line of research performs 2D image editing~\cite{instructpix2pix} and use them to update 3D representations, while the challenge is that per-image editing lacks multi-view consistency. Some efforts have be made to alleviate this issue by introducing noise synchronization~\cite{consistdreamer}, pixel correspondence constraint~\cite{syncnoise, efficientn2n}, or cross-attention mechanism~\cite{dge, trame}, but 
the challenge remains. In addition, these methods heavily rely on iterative 3D updating~\cite{instructn2n, dreamcatalyst}, resulting in slow convergence.
We argue that the common problem of these approaches lies in the lack of inter-frame consistency priors in 2D models. In this work, we propose to explore generative video models for efficient multi-view consistent 3D editing.

\noindent \textbf{Video Generation and Editing}.
Animatediff~\cite{animatediff}, SVD~\cite{svd} and their following works~\cite{vgen,videocrafter1,videocrafter2,sora, opensora, cogvideo,cogvideox}
extend pre-trained 2D image generative model for video generation by fine-tuning them with video datasets and incorporating temporal information interactions~\cite{3dunet,attention}. However, the generated videos are uncontrollable in camera pose. Therefore, 
CameraCtrl~\cite{cameractrl} and MotionCtrl~\cite{motionctrl} introduce extra camera trajectory as input to enable video generation with customized camera motion. 
Tune-a-video~\cite{tuneavideo}, FateZero~\cite{fatezero} and Tokenflow~\cite{tokenflow} are typical works to accomplish video editing with pre-trained SD~\cite{sd} model, yet they can only achieve sub-optimal performance due to the limited capability of 2D models in processing videos. 
MotionI2V~\cite{motioni2v}, MOCA~\cite{moca}, VideoSwap~\cite{videoswap} and I2VEdit~\cite{i2vedit}  employ more powerful video models to perform video editing. However, they demand complex preprocessing of optical flow estimation, key point tracking, or motion LoRA optimization to control the video editing process. 
While MotionClone~\cite{motionclone}, Videoshop~\cite{videoshop},  AnyV2V~\cite{anyv2v}, and VACE~\cite{vace} perform video editing in a training-free or learning-based strategy, they lack 3D geometric constraints and often generate inconsistent results.

\begin{figure*}[ht!]
    \centering
    \includegraphics[width=0.95\linewidth]{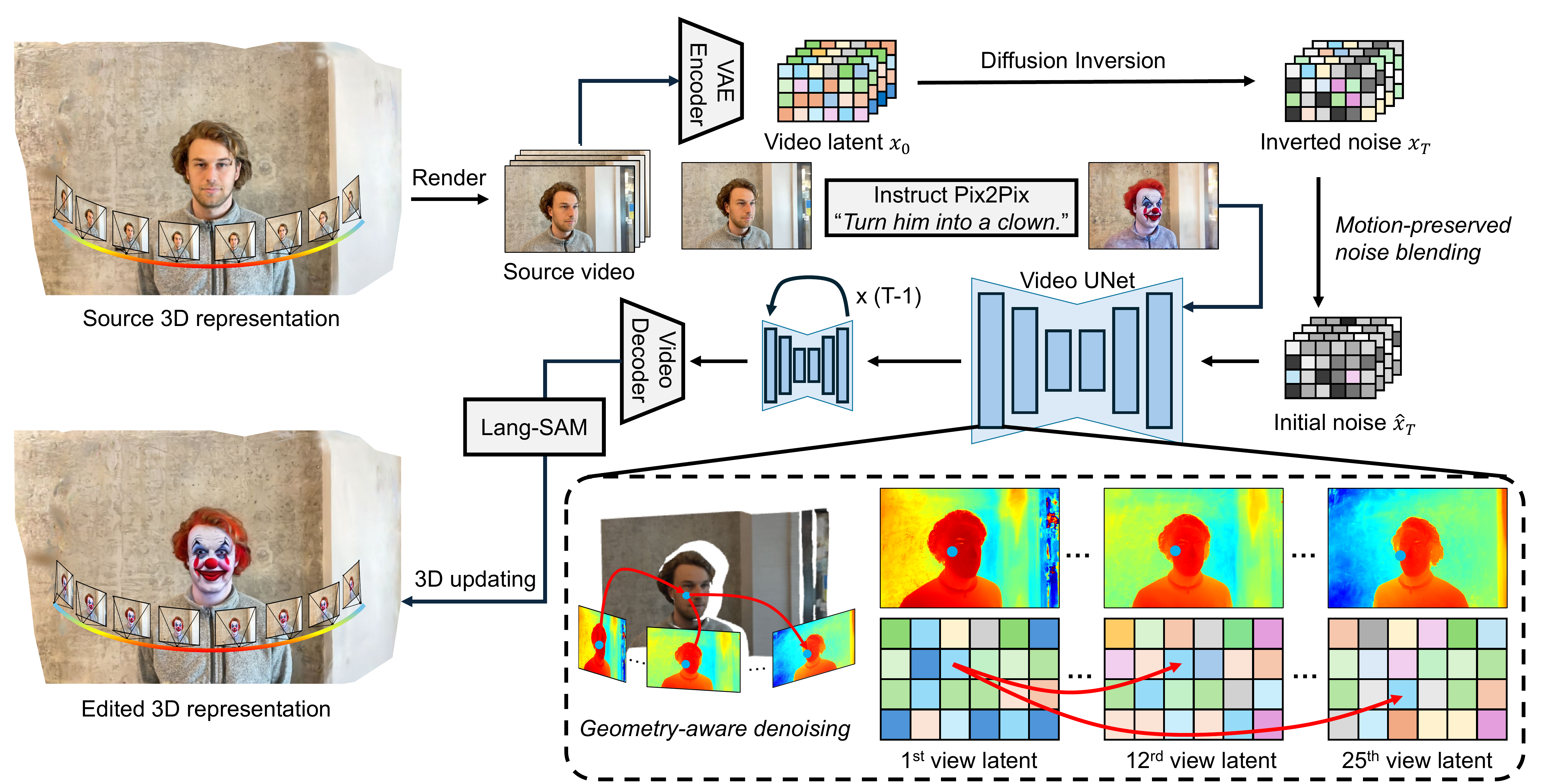}
    \caption{\textbf{Workflow of ViP3DE.} First, the contiguous multi-view images are rendered from the source 3D representation as source video. 
    Then, the first frame is edited with InstructPix2Pix as the condition of the video model, and the initial noise of the diffusion process is obtained by \textit{motion-preserved noise blending}. Consequently, the geometric priors excavated from the source 3D representation are introduced during the video denoising process to improve 3D consistency across views, termed \textit{geometry-aware denoising}.
    Finally, these edited multi-view images are utilized to update the source 3D representation. Thanks to video priors, ViP3DE achieves fast and multi-view consistent 3D editing in a single forward pass.}
    \label{fig:framework}
        \vspace{-6pt}
\end{figure*}

\section{The Proposed Approach}
\label{sec:v3de}

Following common 3D editing protocols, ViP3DE begins with a 3D scene from systems like COLMAP~\cite{colmap} and user-provided instructions. The 3D editing is performed according to the user-provided instructions. 
ViP3DE employs SVD-XT \cite{svd} for its competitive performance with much faster inference speed compared to large models $e.g.$, Wan2.2~\cite{wan22}, CogvideoX~\cite{cogvideox}.
The ViP3DE workflow is illustrated in Fig.~\ref{fig:framework}.
Firstly, we render the 3D scene to obtain source views following continuous camera trajectories, termed source video. 
Then, we propose \textit{motion-preserved noise blending} and \textit{geometry-aware denoising} to achieve geometrically consistent edited multi-views, which are used to update source 3D Gaussians.


\subsection{Editing Multi-view Images with Video Prior}
\label{sec:baseline}

ViP3DE accomplishes multi-view editing by integrating InstructPix2Pix~\cite{instructpix2pix} with SVD-XT in an inversion-based manner~\cite{videoshop, anyv2v, i2vedit}.
In particular, for a source video $\mathcal{I}_{src}=\{I_{src}^1, ..., I_{src}^N\}$ with $N$-view, its latent $x_0$ is first obtained by the VAE encoder, then passed through EDM~\cite{edm} inversion to get the inverted latent noise $x_t$ at step $t=1,...,T$. 
The first frame $I_{src}^1$ and textual instruction are fed into InstructionPix2Pix to obtain an edited view $I_{cond}^1$, which serves as the condition to guide the denoising process, producing edited video latent $\widehat{x}_0$. Finally, $\widehat{x}_0$ is converted to the edited video $\mathcal{I}_{edit}=\{I_{edit}^1, ..., I_{edit}^N\}$ using the pre-trained video decoder. The details of EDM inversion can be found in the  
\textbf{supplementary file}.
We take advantage of the byproduct of EDM inversion ($i.e.$, inverted noise and attention maps) to inherit the camera motion from the source video to output 3D consistent edited views, which can be directly used to update the 3D representation in known camera poses.

\begin{figure}[t]
    \centering
    \begin{minipage}[t]{1\linewidth}
            \includegraphics[width=1\linewidth]{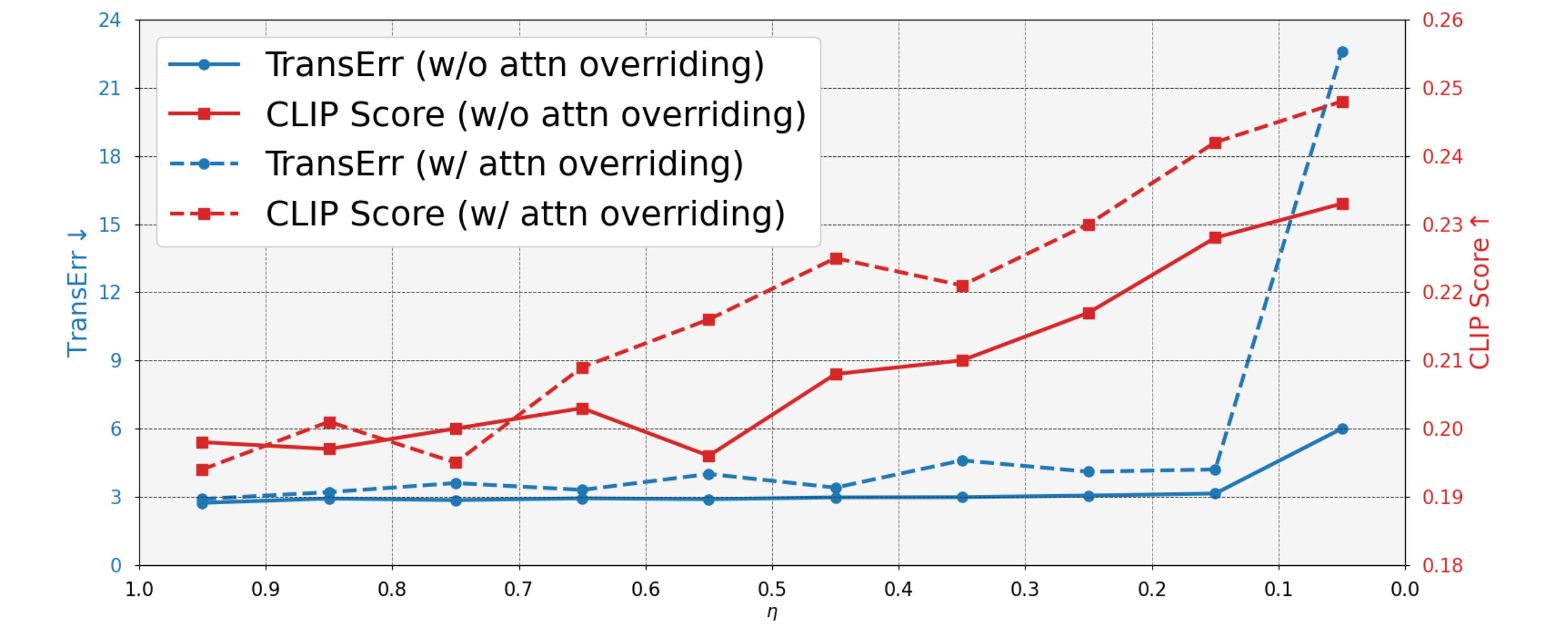}

    \end{minipage}
    
    \vspace{-2mm}
    \small{(a) The values of TransErr and CLIP score versus $\eta$.}

    \begin{minipage}[t]{1\linewidth}
            \includegraphics[width=1\linewidth]{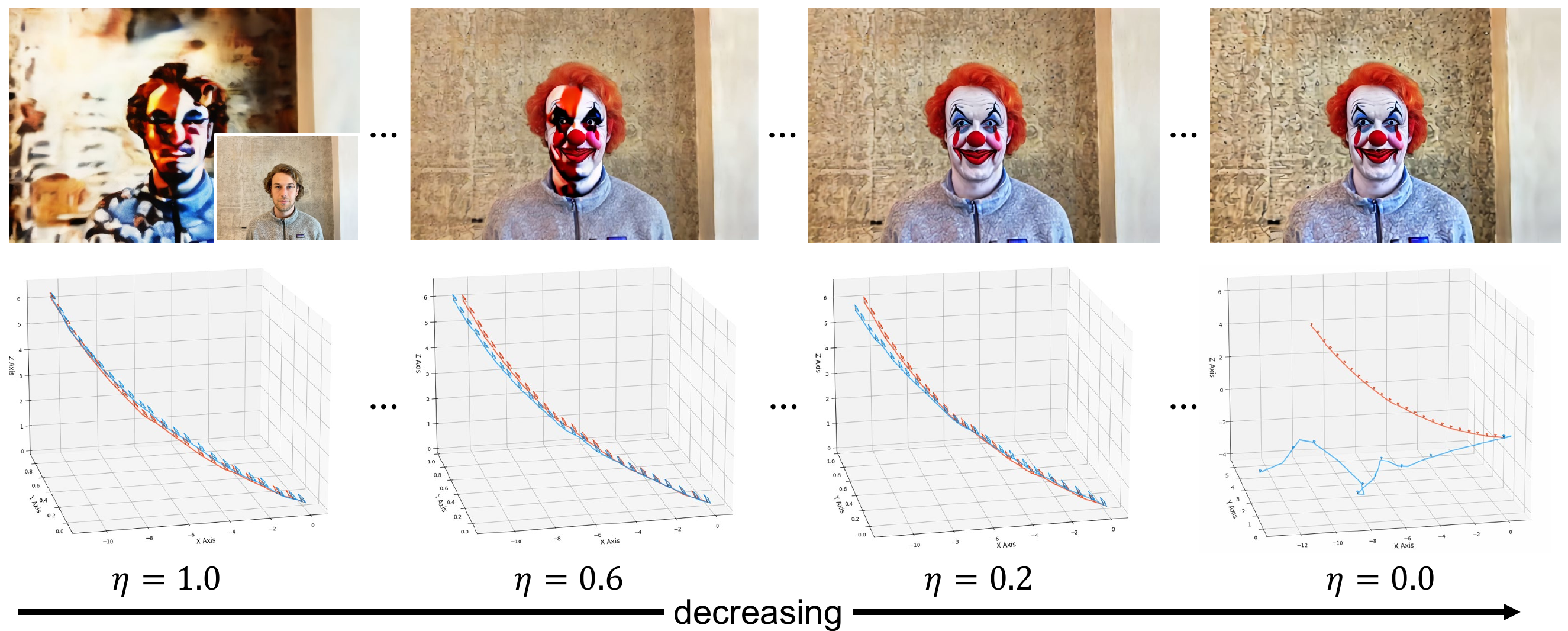}
    \end{minipage}
    
   \vspace{-1mm}
    \small{(b)  Edited first frame and camera poses with different $\eta$.}
    
    \vspace{-2mm}
    \caption{\textbf{Demonstration of motion-preserved noise blending.} Appearance and pose alignment exhibit different levels of robustness to noise.  }
    \label{fig:ablation_in_method}
    
\vspace{-5mm}
\end{figure}

\subsection{Balance Pose Alignment and Editing Quality}
\label{sec:balance}
It is noticed that by starting the generation process from $x_T$, the edited multi-view images can achieve desired camera pose alignment. However, they exhibit significant artifacts if the instruction indicates substantial modifications, $e.g.$, ``\textit{Turn the man into a clown}'', as shown in the first column of Fig.~\ref{fig:ablation_in_method}(b).
We hypothesize that this is because the inverted noise $x_T$ contains not only motion cues but also the appearance cues of the source video.
When conditioned on an edited image $I_{cond}^1$ that significantly deviates from the appearance of source video, the network can be confused and produce ambiguous results. Therefore, there is a critical question: 
\textit{Can we retain the camera pose/motion information while removing unwanted appearance cues in the inverted noise?}

We experimentally find that motion and appearance cues exhibit different behaviors as the initial noise progressively transitions from inverted noise $x_T$ to random Gaussian noise $\epsilon$, enabling the disentanglement of motion cues and appearance cues.
Specifically, we perform a pilot study by randomly selecting 100 off-the-shelf 3D assets~\cite{instructn2n, llff} and render them into 3D videos. Then we leverage GPT-4o~\cite{gpt4o} to generate editing instructions and target prompts based on the first frame of the source video, which are utilized to generate the edited video by noise inversion. The initial noise $\widehat{x}_T$ is obtained by blending the inverted noise $x_T$ and random Gaussian noise $\epsilon$ with different weights: 
\begin{equation}
    \widehat{x}_T = \sqrt{\eta}x_T + \sqrt{1-\eta}\epsilon, \text{where} \ \epsilon \sim \mathcal{N}(0,\sigma_T^2 \textbf{I}),
\label{eq:appearance_reduction}
\end{equation}
where $\eta \in [0, 1]$ controls the intensity of inverted noise. We call this operation \textit{motion-preserved noise blending}.

We employ COLMAP~\cite{colmap} and TransErr~\cite{cameractrl} to estimate and compare the camera poses of edited and source videos with different $\widehat{x}_T$, and use CLIP text-image score~\cite{clip} to study the faithfulness of edited first frame $I_{edit}^1$. We analyze both results with and without \textit{attention overriding}, which is a typical technique to improve camera pose alignment~\cite{anyv2v, motionclone}. The results under different $\eta$ are plotted in Fig.~\ref{fig:ablation_in_method}(a).
One can observe that as $\eta$ gradually decreases from 1 to 0, the quality of edited frames improves steadily, with persistent performance gains within the entire range of $\eta$. Meanwhile, the camera pose alignment exhibits remarkable robustness, maintaining highly competitive \text{TransErr} even when the signal-to-noise ratio is low ($i.e.$, $\eta=0.1$). Note that $\eta=0$ corresponds to the completely free generation with random camera trajectories. 
The editing example ``\textit{Turn the man into a clown}'' is shown in Fig.~\ref{fig:ablation_in_method}(b).

The above findings can be interpreted from two perspectives. Firstly, setting initial noise close to random Gaussian noise allows better appearance manipulation, which is identical to the conclusion in  image editing works~\cite{nulltext,ruibin}.
Secondly, we posit that camera motion constitutes global low-frequency information, making it relatively insensitive to additive white Gaussian noise. More analyses can be found in the \textbf{supplementary file.} Based on the different behaviors between motion and appearance cues, we set $\eta=0.15$ to balance camera poses alignment and editing quality.

\begin{table*}[t]
\vspace{-5pt}
    \centering
    \resizebox{0.9\linewidth}{!}{
    \begin{tabular}{lcccccc}
        \toprule
        \multirow{2}{*}{Method} &  \multirow{1}{*}{Multi-modal } & \multirow{1}{*}{Consistency }  & CLIP & CLIP &   \multirow{1}{*}{Forward }    & \multirow{2}{*}{Time} \\
        & Models & Mechanism &  T-I Sim. & Direction Sim. & Pass  \\
        \hline
        \textit{3D-based methods} \\
        NeRF-Art~\small{~\cite{nerfart}} & CLIP  & \small{Iterative Updating}  & 0.243 & 0.121 & 400 & $>$ 8 hours \\
        ViCA-NeRF~\small{~\cite{vicanerf}} & InstructPix2Pix & \small{Depth Constraint} & 0.274 & 0.183 &  2 & $\sim$ 25 min \\
        GaussCtrl~\small{~\cite{gaussctrl}} & ControlNet & \small{Depth Constraint}  & 0.266 & 0.170   & 1 & $\sim$ 10 min \\
        InstructN2N~\small{~\cite{instructn2n}}   & InstructPix2Pix  & \small{Iterative Updating}    & 0.262 & 0.145  &  5000 &   $\sim$ 28 min \\
        GaussianEditor~\small{\cite{gaussianeditor}}    &   InstructPix2Pix    & \small{Iterative Updating}    & 0.272   & 0.187   &  1500  & $\sim$ 8 min \\
        DGE~\small{\cite{dge}}  &  InstructPix2Pix & \small{Extrapolated Attention}   & 0.269 & 0.180 &  3 & $\sim$ 4 min \\
        \midrule
        \textit{Video-based methods} \\
        AnyV2V~\cite{anyv2v} & I2VGen & Video Editing & 0.230 & 0.114 & 1 & $\sim$12 min \\
        VideoShop~\cite{videoshop} & SVD-XT & Video Editing & 0.253 & 0.149 & 1 & $\sim$ 3 min  \\

        I2VEdit~\cite{i2vedit} & SVD-XT & Video Editing & 0.264 & 0.178 & 1 &  $\sim$ 40 min \\
        InsViE~\cite{wu2025insvie} & CogVideoX & Video Editing & 0.244 & 0.132 & 1 & $\sim$ 4 min \\
        \midrule
        \textbf{ViP3DE (Ours)} &  SVD-XT  & \small{Video\&3D Prior}       &  \textbf{0.284} & \textbf{0.197}  &  1 & $\sim$ 3 min \\
        \bottomrule
    \end{tabular}
    }
    \caption{ \textbf{Quantitative comparison with previous methods.} ViP3DE achieves more faithful results to the texture instruction with higher CLIP scores. In addition, ViP3DE costs less time to converge. The consistency mechanisms of video-based methods are set to `-' since they are not designed for 3D consistent editing.}
        \vspace{-10pt}
    \label{tab:comparison}
\end{table*}

\subsection{Integrating  3D Priors and Video Priors}
Equipped with motion-preserved noise blending, our model can generate high-quality edited views under specific camera poses. However, we observe that parts of edited views suffer from structural deformation and color shifts. This is because while the generated videos exhibit inter-frame continuity, they are not consistent in 3D space. This discrepancy stems from the fact that current generative video models lack a comprehensive understanding of the real-world 3D geometry~\cite{physic,sora}. 
To address this issue, we propose \textit{geometry-aware denoising} to exploit the geometric priors from the source 3D Gaussians to guide the denoising process of video generation.

\begin{algorithm}[t]
\textbf{Input:} Source 3D views $\mathcal{I}_{src}=\{I_{src}^1,..., I_{src}^N\}$, hyper-parameter $\eta$, $\tau$, classifier guidance $w$, geometric constraint index $M$, and instruction $\mathcal{P}$. \\
\textbf{Output:} Edited 3D views  $\mathcal{I}_{edit}=\{{I}_{edit}^1,...,I_{edit}^N\}$.\\
$\{x_1, x_2,...x_T\}\gets \text{EDM-Inv}(\mathcal{I}, I_{src}^1)$    \\
$I_{cond}^1 \gets Edit(I_{src}^1; \mathcal{P})$  \\
$\widehat{x}_T=\sqrt{\alpha}x_T+ \sqrt{1-\alpha}\epsilon, \text{where } \  \epsilon \sim \mathcal{N}(0,\sigma_T^2 \textbf{I})  $ \\ 
\For{$t=T,T-1, \ldots, 1$}{
         $\widehat{x}_{t}^{uc}\gets \widehat{x}_{t}$ \\
         $\widehat{x}_{t}^c\gets \widehat{x}_{t}[M]$ \ \ \# geometry-aware denoising \\
         $\widehat{x}_{t-1}\gets (1+w)\text{GVM}(\widehat{x}_t^{c}, I_{cond}^1; t)-w\text{GVM}(\widehat{x}_t^{uc}, \varnothing; t)$ \ \ \# classifier-free guidance \\
     }
$\mathcal{I}_{edit} \gets \widehat{x}_0$ \\
 \textbf{Return} $\mathcal{I}_{edit}$
 \caption{ 3D Consistent Editing with ViP3DE}
\label{alg:ViP3DE}
\end{algorithm}

We begin by rendering the depth of the source Gaussian to obtain depth maps $\mathcal{D}=\{D_1, ..., D_N\}$ from $N$ views. 
Based on known camera poses, we identify the corresponding pixel at $(\hat{u},\hat{v})$ in the first conditional frame for the pixel at $(u,v)$ in the $i$-th frame through 3D projection:
\begin{equation}
    [\hat{u}, \hat{v},\widehat{D}_{1}(\hat{u}, \hat{v})]^\text{T}    =  \text{K}_{1} \text{Rt}_{1} \text{Rt}_{i}^{-1} \text{K}_{i}^{-1} [u, v, D_{i}(u, v)]^\text{T},
\end{equation}
where $\text{K}_i$ and $\text{Rt}_{i}$ are intrinsic and extrinsic matrices of the $i$-th camera pose, and $\widehat{D}_{1}(\hat{u}, \hat{v})$ is the projected Gaussian depth at $(\hat{u}, \hat{v})$ of the first frame.
Considering that some projected Gaussians may be invisible in the first frame, we filter out occluded pixel correspondences by comparing the projected depth $\widehat{D}_1$ with the rendered depth map $D_1$. The mapping $M$ for $(u, v)$ of the $i$-th frame is defined as:
\begin{equation}
    M(i, u, v)=
    \begin{cases}
       (\hat{u}, \hat{v}) & \text{if}\ |\widehat{D}_1(\hat{u}, \hat{v}) - D_1(\hat{u}, \hat{v})|< \tau, \\
         \varnothing & \text{otherwise},
    \end{cases}
\label{eq:depth}
\end{equation}
where $\tau$ is the threshold to filter occluded correspondences.

The computed geometric correspondence is then leveraged to provide additional guidance during the diffusion denoising process. Specifically, we downsample the pixel correspondences to the latent resolution. 
At each denoising step, the features of all frames in the last layer of the UNet are replaced by those from the corresponding positions in the first frame if valid correspondence exists. 
Furthermore, we exploit the advantage of classifier-free guidance (CFG) by performing feature overriding only on conditional latent features and omitting the unconditional features so that rich texture can be generated from the unconditional prediction term, achieving multi-view consistent 3D editing results.

\noindent \textbf{Pseudo Code.}
Let $Edit(I, \mathcal{P})$ be the image editing function with input image $I$ and editing instruction prompt $\mathcal{P}$. To align with previous work, we employ InstructPix2Pix~\cite{instructpix2pix} to edit the first frame. 
Denote by $GVM(x_t, I; t)$ the computation of a single diffusion step $t$ in the generative video model with noisy video latent $x_t$ conditioned on Image $I$. Let $M$ record the indices of each view's latent features corresponding to the first view latent features, which is used to provide 3D geometric constraints.
The algorithm of ViP3DE is summarized in Alg.~\ref{alg:ViP3DE}.

\noindent \textbf{Remarks.} 
Note that existing methods rely on epipolar constraints~\cite{dge,epidiff} or semantic correspondence~\cite{trame}, which omit 3D Gaussian geometric priors and suffer from inaccurate matching. Studies in~\cite{syncnoise, efficientn2n} enforce pixel-wise constraints, introducing noticeable artifacts and requiring multiple forward passes. 
In contrast, our method addresses these limitations through latent-space integration of 3D priors and video priors, achieving high-fidelity editing in a single forward pass.

\subsection{Implementation Details}
\label{sec:others}

\noindent \textbf{Parallel Inference.}
We perform individual view editing for all video frames and select the one with the highest CLIP similarity score as the condition of SVD-XT. To adapt the first-frame conditional paradigm, we temporally partition the video into two subsequences at the conditional frame while reversing the temporal order of the preceding sub-video. These two sub-videos share the same condition and are edited in parallel to reduce time costs.
An autoregressive manner~\cite{i2vedit} is adopted if the frames of sub-video exceed the model's context length.

\noindent \textbf{Views Continuity.}
We place continuous cameras around the 3D scene to obtain source video. Specifically, we first follow DGE~\cite{dge} to sort the training view cameras according to their view changes. Then, we randomly select several cameras from the sorted training views as the key cameras, which are used to obtain interpolated cameras based on Slerp ~\cite{slerp, quaternions} and linear interpolation. 

\noindent \textbf{Updating 3D Representation.}
We find that InstructPix2Pix tends to override the whole image even if the editing instruction specifies partial localization. Therefore, we follow GaussianEditor~\cite{gaussianeditor} to employ SAM~\cite{sam} to calculate 2D masks, which prevent the 3D updating from occurring in unwanted regions.  Following the typical 3D Gaussians reconstruction~\cite{3dgs}, we update source 3D under the supervision of 2D views with $L_1$ and LPIPS losses~\cite{lpips}.

\begin{figure*}[ht!]
    \centering
    \begin{tabular}{p{0.05\linewidth}p{0.16\linewidth}p{0.1\linewidth}p{0.1\linewidth}p{0.1\linewidth}p{0.12\linewidth}p{0.1\linewidth}}
        {} &
        {\small Reference} & 
        { \scriptsize InstructN2N\cite{instructn2n}} & 
        {\scriptsize ViCANeRF\cite{vicanerf}} & 
        {\scriptsize GaussCtrl\cite{gaussctrl}} & 
        {\scriptsize GaussianEditor\cite{gaussianeditor}} & 
        {\scriptsize ViP3DE} \\
    \end{tabular}
    \vspace{-0.0em} 

    \includegraphics[width=0.9\linewidth]{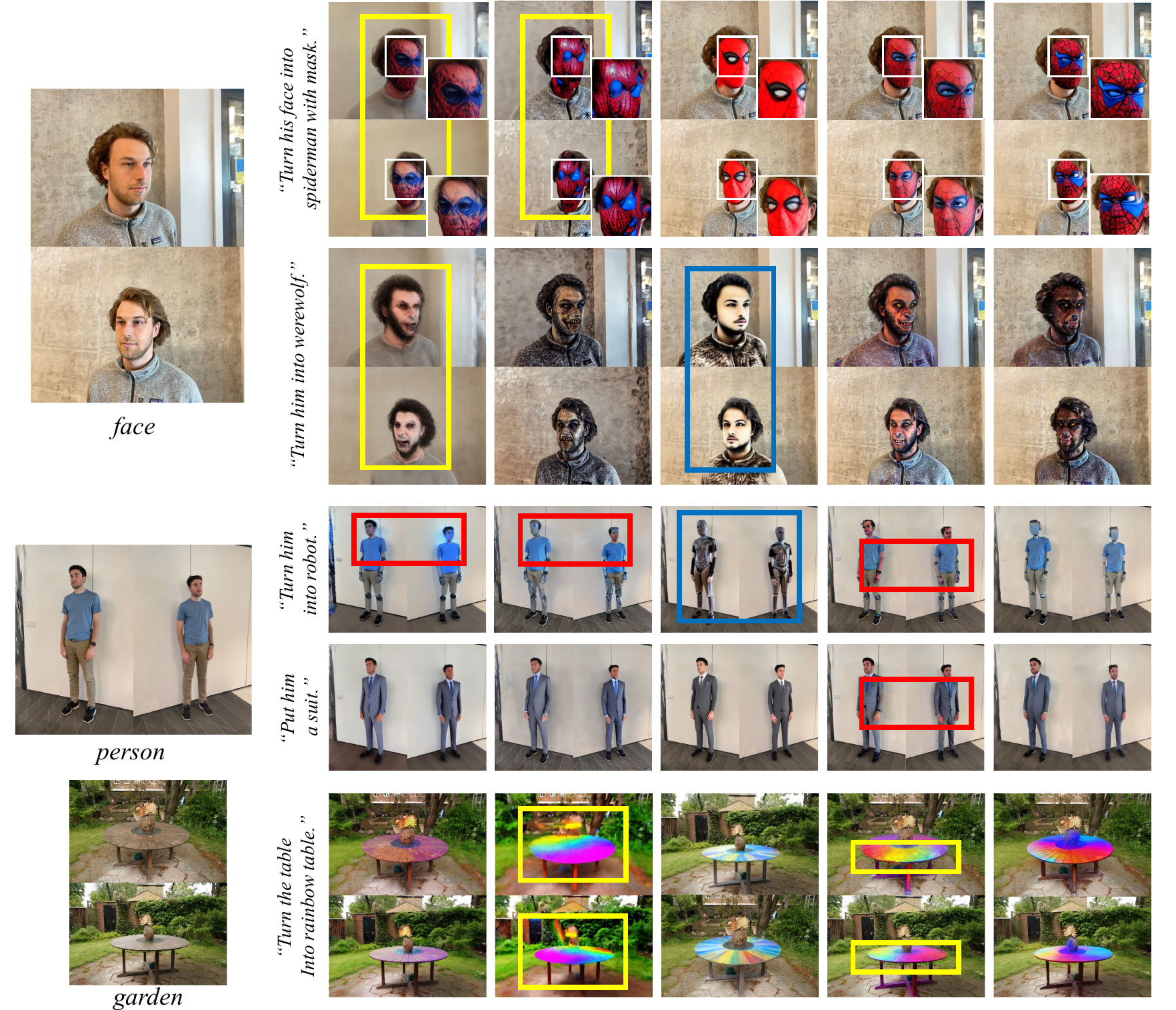}
    \caption{\textbf{Qualitative comparison.} We highlight the edited results suffering from inconsistency (red boxes), poor details (yellow boxes), and unfaithfulness (blue boxes).
    In comparison, ViP3DE obtains consistent results with higher faithfulness to instruction. Besides, rich details are preserved by avoiding multiple iterations that typically cause over-smoothed textures.}
    \label{fig:qualitative_comparison}
    \vspace{-5pt}
\end{figure*}

\section{Experiments}

\textbf{Dataset and Metrics.} To compare ViP3DE with previous methods, we collect 3D scenes and object assets from diverse datasets.  Considering the significant time cost in previous methods like~\cite{nerfart}, we conduct fair comparative evaluations using a subset of Mip-NeRF360~\cite{mipnerf360} and LLFF~\cite{llff}. More details are provided in the \textbf{supplementary file}.
We follow common practice~\cite{instructn2n,dge} using CLIP text-image similarity and directional similarity to evaluate the alignment of editing and instructions, CLIP temporal consistency~\cite{instructn2n} is adopted to study the cross-view consistency.

\noindent \textbf{Experimental Setting.}
The CFGs of textual and image conditions in InstructPix2Pix ~\cite{instructpix2pix} are set to 7.5 and 1.5, respectively.
The numbers of inversion steps and denoising steps in video models are set to 25. 
The $\tau$ in Eq.~\ref{eq:depth} is set to 0.5.
We use edited multi-view images to perform 750 updating iterations.
All experiments are conducted on two RTX A6000 GPUs.

\begin{table*}[t]
\centering
\label{tab:methods_comparison}
\resizebox{.9\linewidth}{!}{
\begin{tabular}{lccccccc}
\toprule
\multirow{2}{*}{Method} & \multirow{2}{*}{Base Model} & \multirow{2}{*}{Video Prior} & \multirow{2}{*}{Training-Free} & \multicolumn{2}{c}{Operations on} &   \multirow{2}{*}{Remark} \\
& & & & Inverted Noise & 3D constraint  & \\
\midrule
Tune-A-Video~\cite{tuneavideo} & SD & \xmark & \xmark & \xmark & \xmark &  No video pre-training \\
TokenFlow~\cite{tokenflow} & SD & \xmark & \cmark & \xmark & \xmark & No video pre-training \\
CoDeF~\cite{codef} & - & \xmark & \xmark & \xmark & \xmark &  Need training, time-consuming  \\ 
ControlVideo~\cite{controlvideo} & ControlNet & \xmark & \cmark & \xmark & \xmark & No video pre-training \\
MotionClone~\cite{motionclone} & AnimateDiff & \cmark & \cmark & \xmark & \xmark & Low editability  \\
AnyV2V~\cite{anyv2v} &  I2VGen &  \cmark & \cmark & \cmark & \xmark & Color shift \\
VideoShop~\cite{videoshop} & SVD-XT & \cmark & \cmark & \cmark & \xmark  & Low conditional faithfulness \\
I2VEdit~\cite{i2vedit} & SVD-XT & \cmark & \xmark & \cmark & \xmark & Need training, time-consuming \\
\hline
Ours & SVD-XT & \cmark  & \cmark & \cmark & \cmark & Consistent 3D editing   \\
\bottomrule
\end{tabular}
}
\caption{\textbf{Comparison with video-based methods.} }
\label{tab:comp_to_video}
\vspace{-7pt}
\end{table*}

\begin{figure}[t]
    \centering
    \includegraphics[width=1\linewidth]{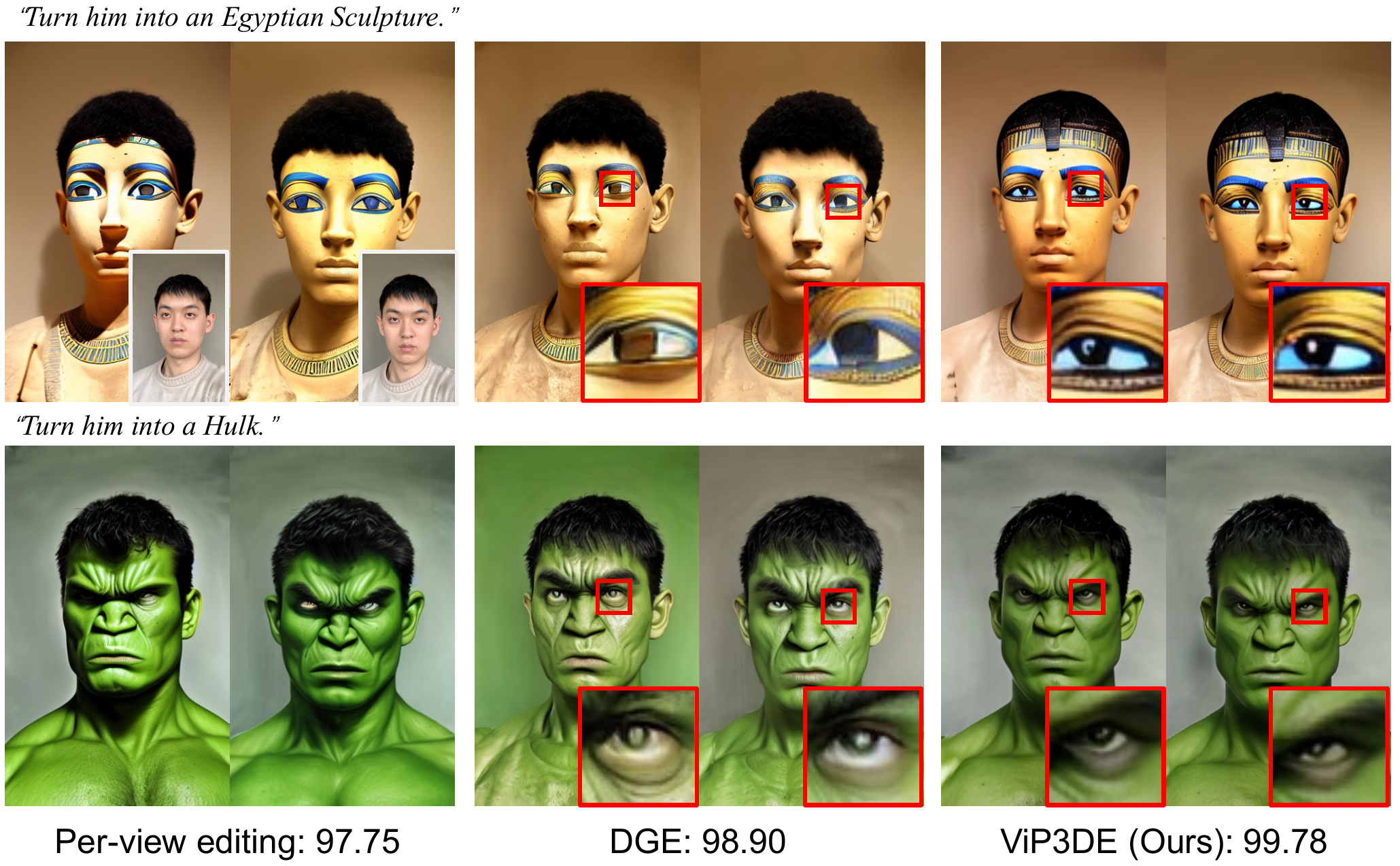}
    \vspace{-5mm}
        \caption{\textbf{The editing results with CLIP temporal score in a single forward pass.} ViP3DE achieves better consistency. }
            \vspace{-1pt}
    \label{fig:comparison_firstphase}
\end{figure}

\subsection{Comparison to 3D Editing Methods}
\noindent \textbf{Quantitative Comparison.} 
We demonstrate the effectiveness and efficiency of ViP3DE by comparing it to previous 3D editing methods.
The numerical results are reported in Tab.~\ref{tab:comparison}. 
Early work NeRF-Art~\cite{nerfart} employs CLIP as a discriminator to edit a 3D object with VolSDF as representation, suffering from low convergence since rendering is time-consuming. 
ViCA-NeRF~\cite{vicanerf} forces the consistency of features in InstructPix2Pix, resulting in over-smoothed editing results with relatively low CLIP scores.
Although GaussianCtrl~\cite{gaussctrl} can achieve 3D editing in a single iteration, it uses ControlNet \cite{controlnet}  to perform editing, which often produces  unfaithful results to the given text prompt. 
InstructN2N~\cite{instructn2n} and GaussianEditor~\cite{gaussianeditor} take advantage of InstructPix2Pix to achieve high-quality editing. However, they edit each view independently, resulting in cross-view inconsistency. In addition, they rely on multiple forward passes, causing slow convergence.
DGE~\cite{dge} introduce semantic correspondence from TokenFLow~\cite{tokenflow} to improve consistency. However, it inherits the limitations of 2D models, lacking cross-view consistency priors.
Instead, ViP3DE introduces video priors with proposed motion-preserved noise blending and geometry-aware denoising, making edited views consistent and faithful.

\noindent \textbf{Qualitative Comparison.} We provide visual comparisons to further illustrate the advantages of ViP3DE.
We mainly study three failure cases including  {inconsistency}, {poor texture details}, and {unfaithfulness}.
As shown in Fig.~\ref{fig:qualitative_comparison}, InstructN2N and GaussianEditor lead to view-inconsistent results due to per-image editing. On \textit{robot}, InstructN2N shows inconsistency in person heads. GaussianEditor maintains different hand colors in different views. In addition, it often introduces floating Gaussian artifacts, as shown in the \textit{garden} scene.
ViCANeRF fails to produce satisfied appearance with over-smoothed textures ($e.g.$, \textit{rainbow table}).
Although GaussCtrl can achieve consistent editing, it suffers from unfaithfulness to the given textural prompt, caused by the use of ControlNet. 
DGE~\cite{dge} extends InstructPix2Pix to a video model, sharing a similar motivation to ours. 
We compare the details of edited frames in a single forward pass in Fig.~\ref{fig:comparison_firstphase}. Both DGE and ViP3DE achieve better temporal consistency compared to per-image editing independently. On \textit{fangzhou}, compared to DGE, which fails to maintain detail consistency and demands iterative updating, ViP3DE achieves better consistency and richer textures by fully exploiting video priors from pre-trained video models.

\begin{figure}[t]
    \centering
    \includegraphics[width=0.95\linewidth]{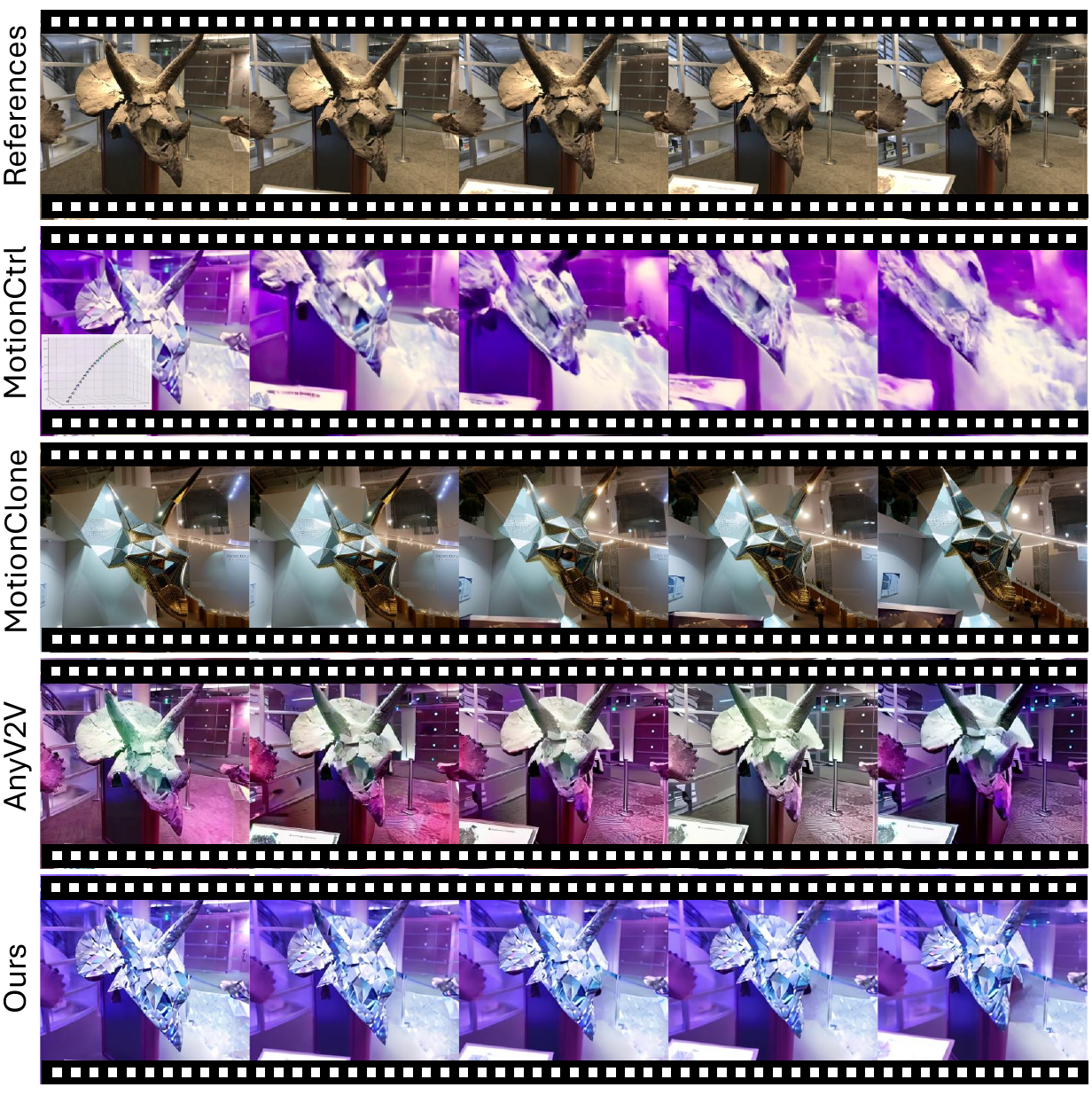}
    \caption{\textbf{Comparison with video editing methods using instruction \textit{``Turn it into crystal diamond horn.''}} ViP3DE achieves better visual quality and camera pose alignment. Note that the images shown here are generated video frames instead of edited 3D Gaussians.}
    \vspace{-10pt}
    \label{fig:video_metho_comp}
\end{figure}

\begin{table}[t]
    \centering
    \resizebox{1\linewidth}{!}{
    \begin{tabular}{ccc|cc}
        \toprule
        Continuous    &   Mition-preserved  & Geometrically aware  &  \multicolumn{2}{c}{CLIP score}  \\
          views &  noise blending & denoising & T-I  & Direction   \\
        \midrule
            &   & &  0.198 & 0.083   \\
          \checkmark  &   &   &  0.220 & 0.109   \\
          \checkmark  & \checkmark   &   & 0.271 & 0.185  \\
          \checkmark  &  & \checkmark   &   0.258 & {0.181} \\
          
          \checkmark  & \checkmark  & \checkmark  & \textbf{0.284} & \textbf{0.197} \\
          
        \bottomrule
    \end{tabular}
    }
    
    \vspace{-4pt}
    \caption{\textbf{Ablation study on ViP3DE.} Removing either motion-preserved noise blending or geometry-aware denoising reduces the performance significantly. }
    \label{tab:ablation}
    \vspace{-12pt}
\end{table}

\begin{figure}[t]
    \centering
    \includegraphics[width=1\linewidth]{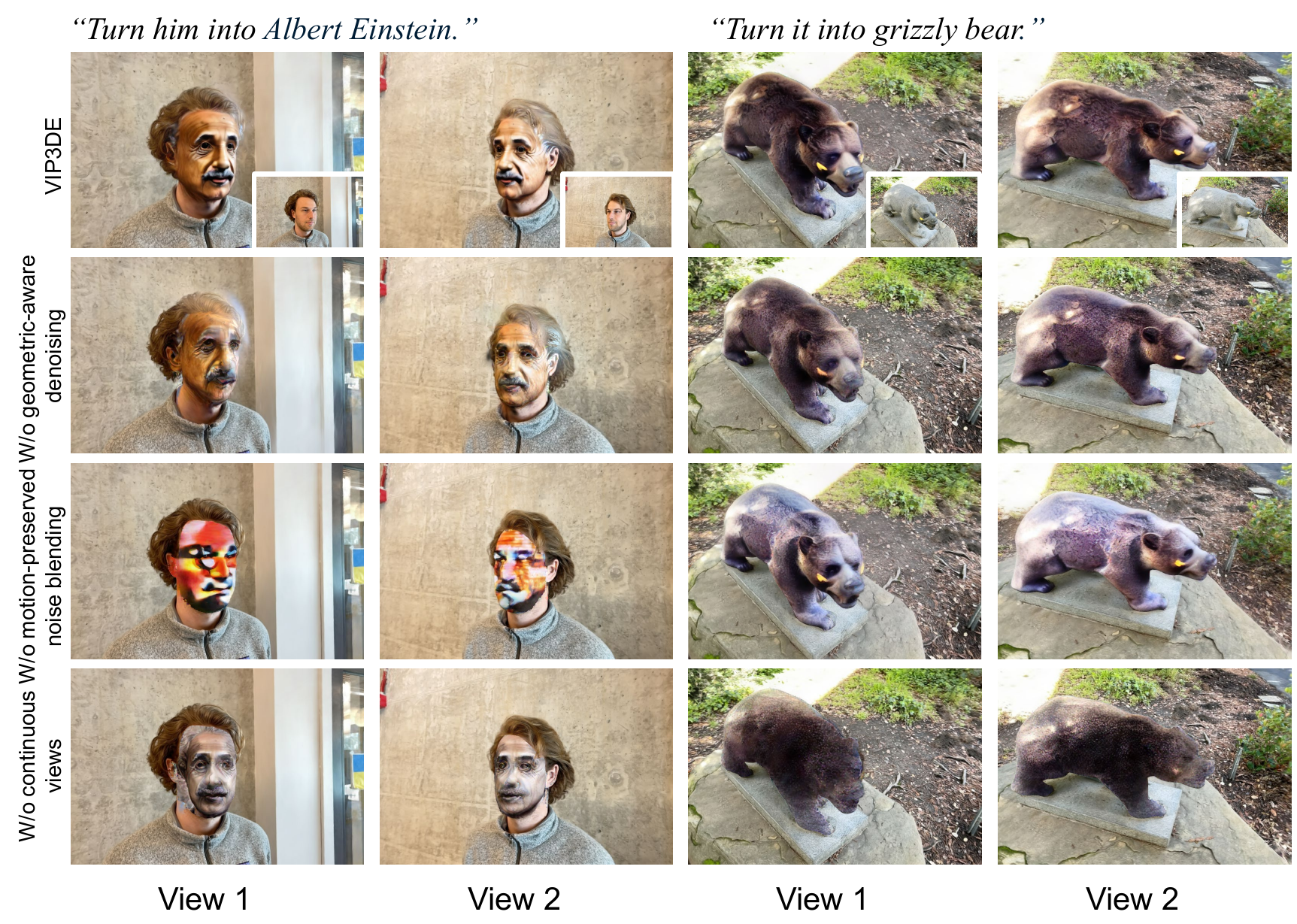}
        \vspace{-7mm}
        \caption{\textbf{Qualitative ablation results.} All the three components are important to achieve 3D consistent editing.  } 
    \label{fig:geometric_aware}
        \vspace{-4pt}
\end{figure}

\subsection{Comparison to Video Editing Methods}
\noindent \textbf{Quantitative Comparison.} 
We compare our method with video-based methods, which has the potential to achieve 3D editing. Both AnyV2V~\cite{anyv2v} and VideoShop~\cite{videoshop} achieve video editing in an inverse-based manner, while InsViE~\cite{wu2025insvie} is an instruction-based video editing model. We employ these methods to edit rendered 3D views in an auto-regressive way and back-project edited views to 3D assets, similar to our post-processing. As shown in Tab.~\ref{tab:comparison}, these methods fail to achieve the desired performance because they overlook the balance between pose alignment and editing quality. Besides, they are designed for general videos editing containing object-level dynamics, which will introduce multi-view inconsistencies when directly applied to 3D editing. In contrast, our proposed ViP3DE is specialized for 3D editing tasks by explicitly integrating 3D geometric priors during editing, thereby enforcing cross-view consistency through physically grounded constraints.
More detailed analysis by comparing ViP3DE with existing video editing methods in Tab.~\ref{tab:comp_to_video}.

\noindent \textbf{Qualitative Comparison.} 
We provide qualitative comparisons with those video-based methods that can be adopted to achieve 3D editing. (1) \textit{Motion-conditioned Video Generation Methods.}
MotionCtrl~\cite{motionctrl} and CammeraCtrl~\cite{cameractrl} aim to generate videos under specified camera poses. 
Specifically, we employ extrinsic matrices of camera poses as the condition, guiding video generation from the edited first frame. 
However, they cannot handle customized trajectories and will produce collapsed content. As shown in the second row of Fig.~\ref{fig:video_metho_comp}, the \textit{crystal horn} with MotionCtrl gradually deviates from its original structure, producing meaningless appearances.
(2) \textit{Video Editing Methods.}  
Recent advances in video editing can also be used to perform 3D editing.
Here, we omit methods that require test-time training ($e.g.$, Tune-a-Video~\cite{tuneavideo}, CoDef~\cite{codef}, and I2VEdit~\cite{i2vedit}), and we compare our ViP3DE with MotionClone~\cite{motionclone} and AnyV2V~\cite{anyv2v} in Fig.~\ref{fig:video_metho_comp}. MotionClone uses a text-guided video model, tending to generate ambiguous edited frames.
The videos generated by AnyV2V show inconsistent colors in the horn. In comparison, ViP3DE achieves better visual quality since we inject 3D priors to enhance consistency.

\subsection{Diagnostic Experiments}
\label{sec:diagnostic}

\noindent \textbf{Quantitative Ablation Studies.}
We conduct ablation studies to evaluate the effectiveness of the key components of ViP3DE in Tab.~\ref{tab:ablation}. We see that the generative video model fails to tackle discrete 3D views and only achieves a CLIP score of 0.198. 
With continuous views as input, editing 3D multi-views with InstructPix2Pix and generative video models still outputs unfaithful visual results. Our proposed motion-preserved noise blending brings a significant performance improvement. Introducing geometry-aware denoising further enforces cross-view 3D consistency, achieving the best performance.

\noindent \textbf{Qualitative Ablation Studies.}
We employ face and bear as examples to demonstrate the functionality of each component. The absence of ordered continuous views in input would make the video model fail in proper view editing, resulting in a mismatch between edited views and their corresponding camera poses. Consequently, the edited 3D representation exhibits significant quality degradation even with mask constraints. Removing motion-preserved noise blending makes the editing unfaithful to the conditional images, showing confused appearance artifacts. In addition, geometry-aware denoising reinforces the geometric relationships among views while maintaining alignment with the source 3D structure, effectively preventing blurring artifacts in the edited 3D views.

\begin{figure}[t]
    \centering
    \includegraphics[width=1\linewidth]{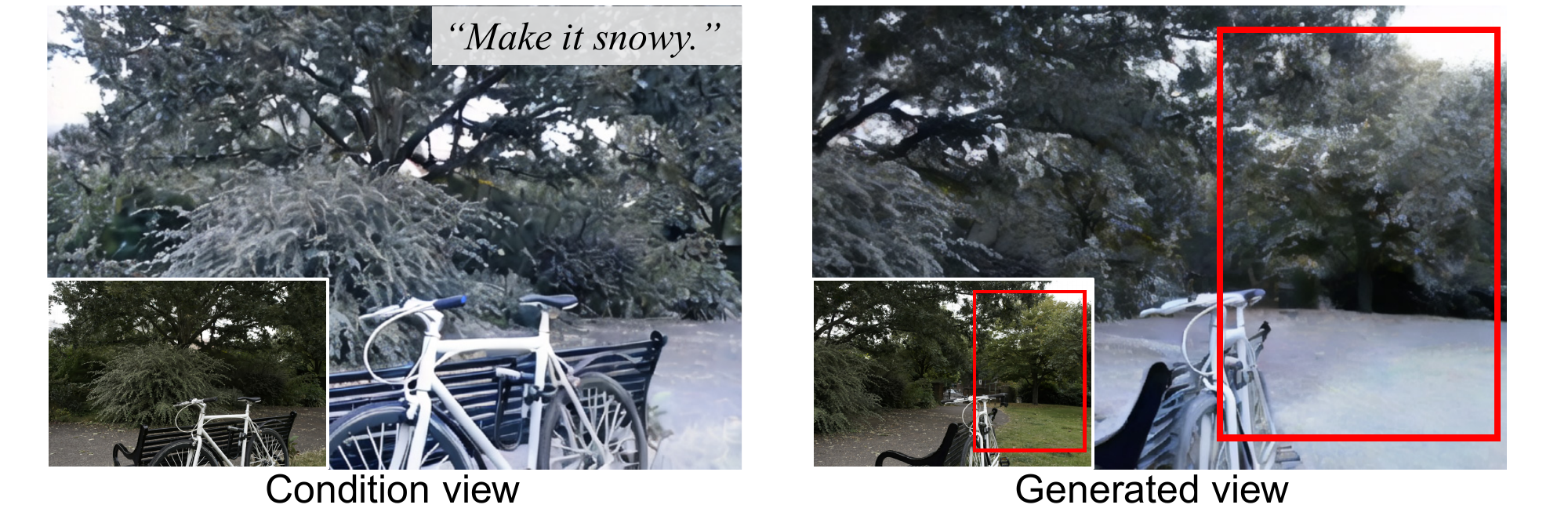}
        \vspace{-7mm}
        \caption{\textbf{The Generalization ability of ViP3DE.} 
        Leveraging video priors, ViP3DE can implicitly transfer the editing cues from the conditional first view to unseen regions. } 
    \label{fig:generation_ability}
        \vspace{-6mm}
\end{figure}

\noindent \textbf{Generation Ability.}
Sometimes, the 3D content to be edited may not be fully covered by the first anchor view. One natural question rises: \textit{Can consistent editing be achieved for missing content?} We use the large \textit{bicycle} scene to demonstrate that ViP3DE inherits the generative capability of the video model and can transfer the editing cues of the anchor view to subsequent content, even if they do not appear in the first frame.
As shown in Fig.~\ref{fig:generation_ability}, the trees and roads within the red box are also covered in snow. 
Note that such a capability is unattainable by previous warping-based single-view editing for 3D editing~\cite{syncnoise}.

We provide more experiments ($e.g.$, comparison with video editing methods), implementation details ($e.g.,$ pilot study), and discussions in the \textbf{supplementary file}.

\section{Conclusion}




In this paper, we proposed ViP3DE by introducing generative video priors for fast and multi-view consistent 3D editing. 
We first rendered continuous views to bridge the gap between discrete 3D views and video. Then, we proposed motion-preserved noise blending to improve editing quality, and introduced geometry-aware denoising to integrate 3D priors with video priors to enhance cross-view 3D consistency. The edited results were used directly to update 3D without iterative passes.
Extensive experimental results demonstrated the superiority of ViP3DE over previous methods in both effectiveness and efficiency.

\noindent \textbf{Limitations.}
While ViP3DE can manage 3D editing with certain geometric changes, such as adding glasses to a person, it is limited in editing scenes with significant geometric alterations, such as raising a person's hands. This limitation mainly inherits from InstructPix2Pix, which is used to generate the edited first frame. As the editing and generative capabilities of image and video models continue to evolve, 3D editing with substantial geometric changes will be solved.


\clearpage

\twocolumn[
  \begin{center}
    \LARGE \textbf{Supplementary Material}
    \vspace{0.5cm} 
  \end{center}
]

\definecolor{my_yellow}{HTML}{FFC500}

\DeclareRobustCommand\onedot{\futurelet\@let@token\@onedot}
\def\@onedot{\ifx\@let@token.\else.\null\fi\xspace}
\def\eg{\emph{e.g}\onedot} \def\Eg{\emph{E.g}\onedot}
\def\ie{\emph{i.e}\onedot} \def\Ie{\emph{I.e}\onedot}

\setcounter{secnumdepth}{1} 

%





\noindent In this supplementary file, we provide the following materials to support the findings of the main paper:

\begin{itemize}
    \item  \textbf{Additional Implementation Details} provide more implementation details of ViP3DE.
    \item \textbf{Additional Analyses} provide extend analyses and discussions of the proposed ViP3DE.
    \item \textbf{Additional Experimental Results} provide more experimental details and visualization.
\end{itemize}

\section{Additional Implementation Details}
\label{sec:method_detail}
\begin{figure*}[h]
    \centering
    \includegraphics[width=0.9\linewidth]{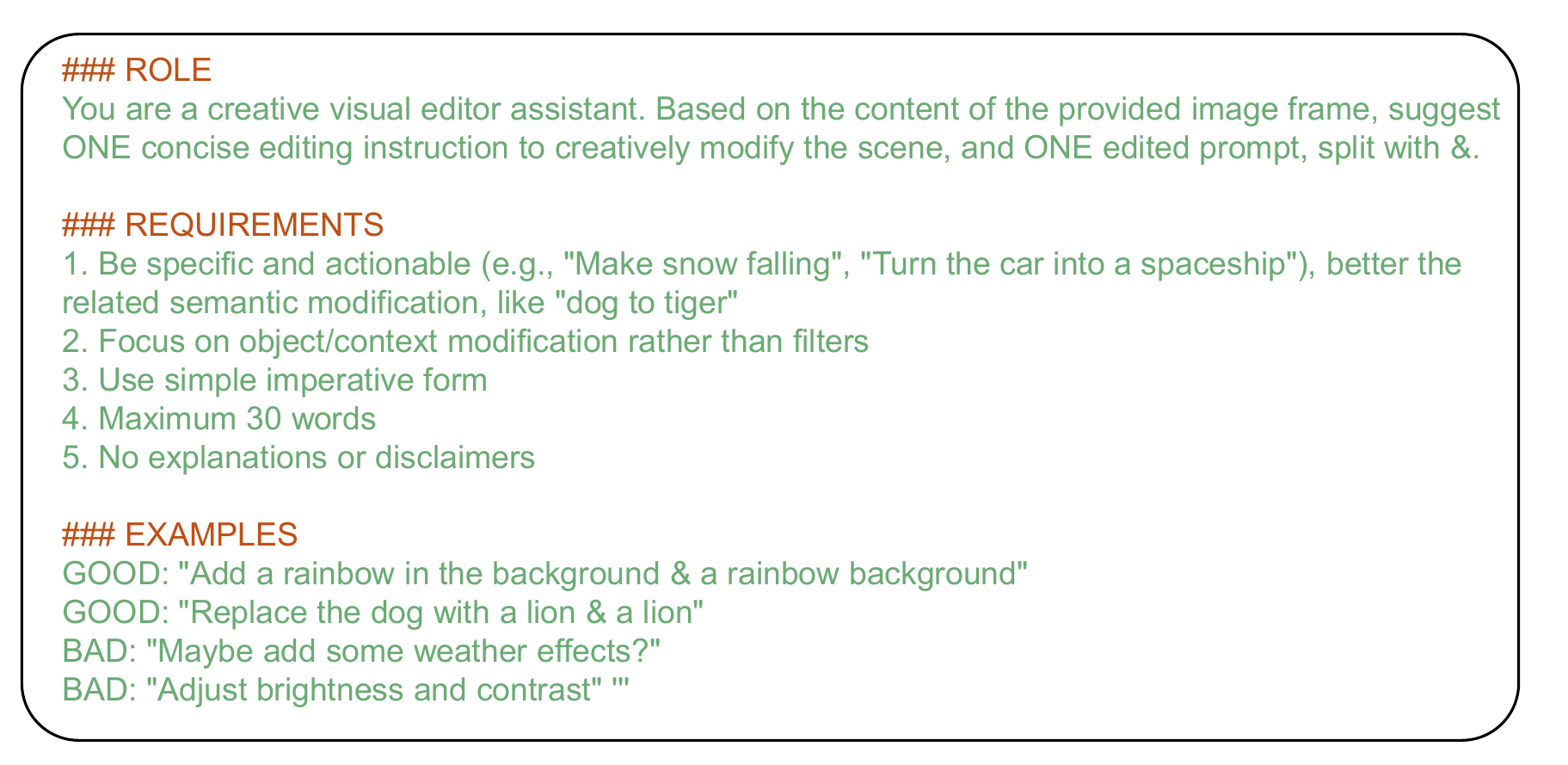}
        \caption{\textbf{The prompt for GPT4o to generate instruction and target prompt automatically.} } 
    \label{fig:suppl_prompt}
\end{figure*}

\noindent \textbf{Generating 3D Instruction with GPT4o.}
As described in the main paper, we curate 100 3D assets and employ GPT4o~\cite{gpt4o} to automatically generate editing instructions and target prompts based on rendered views. The prompt provided to GPT4o is shown in Fig.~\ref{fig:suppl_prompt}.

\noindent \textbf{EDM Inversion.} As generative image-to-video models, SVD ~\cite{svd} and its extension SVD-XT can generate 16 and 25 continuous frames with a given first-frame image as the condition. In the training stage, the video model grounds the initial model on Stable Diffusion~\cite{sd}, followed by video fine-tuning. In the inference stage, SVD-XT denoises the randomly sampled noise $x_T$ with the EDM solver~\cite{edm} as follows\footnote{For simplicity, we omit the $2^{\text{nd}}$ order correction. The noise level increases with the step $t$, which is opposite to the original paper~\cite{edm}. }:
\begin{equation}
    x_t = x_{t+1}+\frac{\sigma_t-\sigma_{t+1}}{\sigma_{t+1}}(x_{t+1}-D_{\theta}(x_{t+1}, t+1; I)) ,
\end{equation}
where $\sigma_t$ is the scheduled noise level at step $t \in [0, T]$. $D_{\theta}(x_t,t;I)$ is the estimated $x_0$ from $x_t$ conditioned on the first frame $I$, which is defined as follows: 
\begin{equation}
    D_{\theta}(x_t, t; I) = x_{t}-(c_{skip}^{t}x_{t} + c_{out}^{t}F_{\theta}(c_{in}^{t}x_{t}, c_{noise}^{t}; I) ) ,
\label{eq:edm}
\end{equation}
where  $c_{skip}^{t}$, $c_{in}^{t}$, $c_{out}^{t}$, and $c_{noise}^{t}$ are coefficients of noise schedule in EDM. $F_{\theta}$ refers to the UNet in SVD framework. SVD extends the classifier-free guidance~\cite{classifierfree} by linearly increasing the guidance scale across the frame axis.
As an ODE approximation solver, EDM significantly reduces the sampling steps, improving the ViP3DE inference speed as well.

 To inverse the denosing diffusion process, Equation~(\ref{eq:edm}) can be rewritten as follows to obtain ${x}_{t+1}$ from ${x}_{t}$:
\begin{equation}
    x_{t+1} = \frac{\sigma_{t+1}x_t+(\sigma_{t}-\sigma_{t+1})c_{out}^{t+1}F(c_{in}^{t+1}x_{t+1}, c_{noise}^{t+1}; I)}{\sigma_{t}-\sigma_{t}c_{skip}^{t+1}+\sigma_{t+1}c_{skip}^{t+1}} .
\end{equation}
Since $x_{t+1}$ is not available during inversion, we approximate $F_{\theta}(c_{in}^{t+1}x_{t+1};c_{noise}^{t+1})$ by employing 
  $F_{\theta} (c_{in}^{t}x_{t};c_{noise}^{t})$ as a substitute.
The inverted noise $x_T$ can be used for source video reconstruction with $I_{src}^1$ as the denoising condition.


\noindent \textbf{Discrete Views Interpolation.}
We first follow DGE~\cite{dge} to sort the training view cameras according to their view changes. Then we randomly select 2$\sim$4 cameras from the sorted training views as the key cameras to be interpolated. 
To obtain a smooth video from a 3D scene, we propose to interpolate two adjacent cameras if they are not close enough. Specifically, given two camera poses $A=\{R_A, T_A\}$ and $B=\{R_B, T_B\}$, their rotations matrices are converted to quaternio $q_A$ and $q_B$ so that linear interpolation can be performed in this space. The quaternion of novel interpolated camera $q_C$ is obtained based on Slerp (spherical linear interpolation)~\cite{slerp, quaternions}, which is formulated as follows: 
\begin{equation}
q_C = \text{Slerp}(q_A, q_B; k) = \frac{\sin \theta}{\sin \theta} q_A + \frac{\sin 
 (1-k)\theta}{\sin \theta} q_B,
\end{equation}
where $\theta=\text{arccos} (q_A\cdot q_B)$, $k \in [0, 1]$ is the interpolation ratio. For translation terms, $T_C$ are interpolated from:
\begin{equation}
T_C = kT_A + (1-k)T_B.
\end{equation}

A number of $K$ interpolated cameras between $A$ and $B$ can be obtained by set $k=\{\frac{1}{K+1}, \frac{2}{K+1}..., \frac{K}{K+1} \}$.

\begin{figure*}[t]
    \centering
    \includegraphics[width=0.9\linewidth]{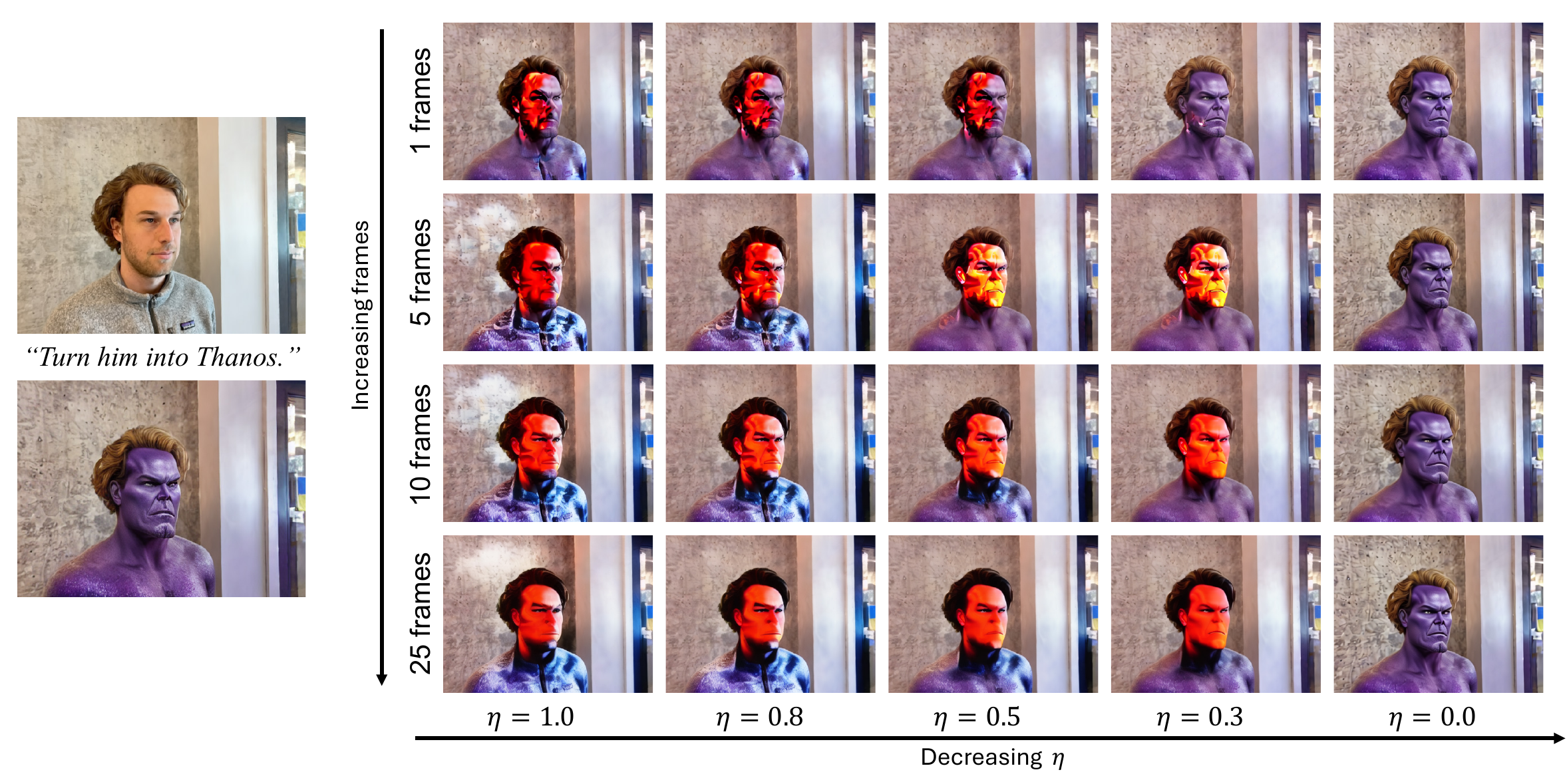}
    \vspace{-5pt}
        \caption{\textbf{The effect of number of views on editing.} } 
    \label{fig:suppl_redundancy}
\end{figure*}

\begin{table*}[h!]
\centering
\resizebox{0.9\linewidth}{!}{
\begin{tabular}{@{}lllll@{}}
\toprule
\textbf{Scene} & \textbf{Source Prompt} & \textbf{Target Prompt} & \textbf{Edit Instruction} & \textbf{Prompt for GaussCtrl} \\ \midrule
Face & \begin{tabular}[c]{@{}l@{}}``A man.''\end{tabular} & \begin{tabular}[c]{@{}l@{}}``A clown.''\end{tabular} & \begin{tabular}[c]{@{}l@{}}``Turn him into a clown.''\end{tabular} & \begin{tabular}[c]{@{}l@{}}``A photo of a man.'' \\ ``A photo of a clown.''\end{tabular} \\
Face & \begin{tabular}[c]{@{}l@{}}``A man.''\end{tabular} & \begin{tabular}[c]{@{}l@{}}``A spiderman with mask.''\end{tabular} & \begin{tabular}[c]{@{}l@{}}``Turn him into a spider  \\ man with mask.''\end{tabular} & \begin{tabular}[c]{@{}l@{}}``A photo of a man.'' \\ ``A photo of a spider man with mask.''\end{tabular} \\
Face & \begin{tabular}[c]{@{}l@{}}``A man.''\end{tabular} & \begin{tabular}[c]{@{}l@{}}``A werewolf.''\end{tabular} & \begin{tabular}[c]{@{}l@{}}``Turn him into a werewolf.''\end{tabular}  & \begin{tabular}[c]{@{}l@{}}``A photo of a man.'' \\ ``A photo of a werewolf.''\end{tabular}  \\

Person & \begin{tabular}[c]{@{}l@{}}``A person.''\end{tabular} & \begin{tabular}[c]{@{}l@{}}``A robot.''\end{tabular} & \begin{tabular}[c]{@{}l@{}}``Turn him into a robot.''\end{tabular} & \begin{tabular}[c]{@{}l@{}}``a photo of a person.'' \\ ``A photo of a robot.''\end{tabular} \\
Person & \begin{tabular}[c]{@{}l@{}}``A person.''\end{tabular} & \begin{tabular}[c]{@{}l@{}}``A person wearing a suit.''\end{tabular} & \begin{tabular}[c]{@{}l@{}}``Put him a suit.''\end{tabular} & \begin{tabular}[c]{@{}l@{}}``A photo of a man.'' \\ ``A photo of a man wearing a suit.''\end{tabular}\\

Garden & \begin{tabular}[c]{@{}l@{}}``A table in the garden.''\end{tabular} & \begin{tabular}[c]{@{}l@{}}``A rainbow table in the garden.''\end{tabular} & \begin{tabular}[c]{@{}l@{}}``Turn the table into a \\ rainbow table.''\end{tabular} & \begin{tabular}[c]{@{}l@{}}``A photo of a table.'' \\ ``A photo of a rainbow table.''\end{tabular} \\
Garden & \begin{tabular}[c]{@{}l@{}}``A table in the garden.''\end{tabular} & \begin{tabular}[c]{@{}l@{}}``A table in the garden with \\ van Gogh style.''\end{tabular} & \begin{tabular}[c]{@{}l@{}}``Make it van Gogh style.''\end{tabular} & \begin{tabular}[c]{@{}l@{}}``A photo of a garden.'' \\ ``A photo of garden  with van Gogh style.''\end{tabular} \\

Bear & \begin{tabular}[c]{@{}l@{}}``A stone bear.''\end{tabular} & \begin{tabular}[c]{@{}l@{}}``A grizzly bear.''\end{tabular} & \begin{tabular}[c]{@{}l@{}}``Turn it into a grizzly bear.''\end{tabular} & \begin{tabular}[c]{@{}l@{}}``A photo of a bear.'' \\ ``A photo of a grizzly bear.''\end{tabular}\\

Fangzhou & \begin{tabular}[c]{@{}l@{}}``A man.''\end{tabular} & \begin{tabular}[c]{@{}l@{}}``A Hulk.''\end{tabular} & \begin{tabular}[c]{@{}l@{}}``Turn him into a Hulk.''\end{tabular} &   \begin{tabular}[c]{@{}l@{}}``A photo of a man.'' \\ ``A photo of a Egyptian Sculpture.''\end{tabular} \\
Fangzhou & \begin{tabular}[c]{@{}l@{}}``A man.''\end{tabular} & \begin{tabular}[c]{@{}l@{}}``An Egyptian Sculpture.''\end{tabular} & \begin{tabular}[c]{@{}l@{}}``Make him look like an \\ Egyptian Sculpture.''\end{tabular} & \begin{tabular}[c]{@{}l@{}}``A photo of a man.'' \\ ``A photo of an Egyptian Sculpture.''\end{tabular} \\

 \bottomrule
\end{tabular}
}
\caption{\textbf{The scene and instruction prompts in the datasets.}}
\label{tab:prompts}
\end{table*}

\section{Additional Analyses}
\label{sec:analysis}

\noindent \textbf{Analyses on Inverted Noise.}
While inverted noise can serve as a validated initialization in image editing, our analysis reveals its limitations in 3D editing, where it produces severe artifacts. This stems from the multi-view nature of 3D content: Static objects' visual patterns show strong cross-view similarity, creating redundancy that survives noise inversion. We devise an experiment to quantify the relationship between 3D view counts and inversion efficacy. In detail, we perform naive inversion-based video editing with different numbers of source frames $N=\{1, 5, 10, 25\}$. As shown in Fig.~\ref{fig:suppl_redundancy}, when there is only a single frame, the editability of inverted noise is relatively high, while as the number of frames increases, the appearance of the edited video becomes less similar to the condition. This phenomenon is observed across different $\eta$, demonstrating that the number of 3D views affects redundancy, and more 3D views reduce the editability of inverted noise.

\noindent \textbf{Analyses on Geometry-aware Denoising.}
The proposed geometry-aware denoising operates within the video latent space across all denoising iterations. Its functional mechanisms can be decomposed into two aspects:
(1) The downsampled geometric information integrated into low-resolution latent space establishes soft yet robust constraints. Unlike pixel-space warping that inevitably introduces artifacts~\cite{syncnoise}, this approach enables minor geometric modifications while maintaining coherence, as evidenced by the distinct geometric profiles of ``Fangzhou" and ``Hulk" in the main paper.
(2) Overriding the latent features based on 3D view geometric relationships essentially performs geometric prior-guided annealing on denoised latent representations, which subsequently participates in the iterative diffusion optimization process, sharing a similar mechanism with SDE solvers like DDPM~\cite{ddpm}.


\noindent \textbf{Efficiency.}
To establish connections between different views, previous methods~\cite{dge,consistdreamer, syncnoise} typically extrapolate the attention layers of 2D models ($e.g.$, Stable Diffusion) to maintain view consistency. Since 2D models process different views independently, these methods increase memory cost significantly, thus they can only tackle a small number of views in a single forward. 
In contrast, ViP3DE is based on a video model that stacks different views along the channel dimension. With efficient convolution and decoupled spatial-temporal attention, our method achieves consistent editing results across more views using video priors.

\noindent \textbf{Object Motion.}
The video model can generate videos containing both camera motion and object motion. 
ViP3DE generates edited 3D views with only camera motion via motion-preserved noise blending,  which potentially provides a solution for decoupling camera motion and object motion. 
On the other hand, better utilization of object motion could provide a technical foundation for 4D perception~\cite{gga,voxelmamba} or 3D animation~\cite{wan22}, which we leave for future work.

\section{Additional Experimental Results}
\label{sec:experimental_detail}
\noindent \textbf{Scene-Prompt Pairs.}
We provide the detailed scene-prompt pairs in Tab.~\ref{tab:prompts}. Since GaussCtrl~\cite{gaussctrl} relies on source and target prompts to achieve editing instead of editing instruction, which is different from other methods, we additionally report the prompts used in GaussCtrl. 

\noindent \textbf{Metric.} We employ $\te$ to study the camera pose alignment of the source and edited 3D views. In detail,  COLMAP~\cite{colmap} is utilized to estimate the camera pose sequence of edited videos, which consist of rotation matrixes $\mathbf{R}_{edit}\in \mathbb{R}^{n\times 3 \times 3}$ and translation vector $\mathbf{T}_{edit}\in \mathbb{R}^{n\times 3 \times 1}$. The translation error $\te$ is calculated by comparing with ground truth matrix $\textbf{R}_{gt}$ and $\textbf{T}_{gt}$ as follows:
\begin{equation}
    \te = \sum_{j=1}^n \norm{{ \mathbf{T}_{gt}^j- \mathbf{T}_{edit}^j }}_2.
\end{equation}

Considering the randomness in the world coordinate origin and intrinsic parameters when estimating camera poses with COLMAP, directly comparing camera trajectories of source and edited videos would introduce significant errors. Therefore, we first fix their intrinsic parameters, including the focal length and image coordinate origin. Then, the extrinsic matrices are aligned to the coordinates of the first view before making a comparison. 

\begin{figure*}[h]
    \centering
    \includegraphics[width=0.9\linewidth]{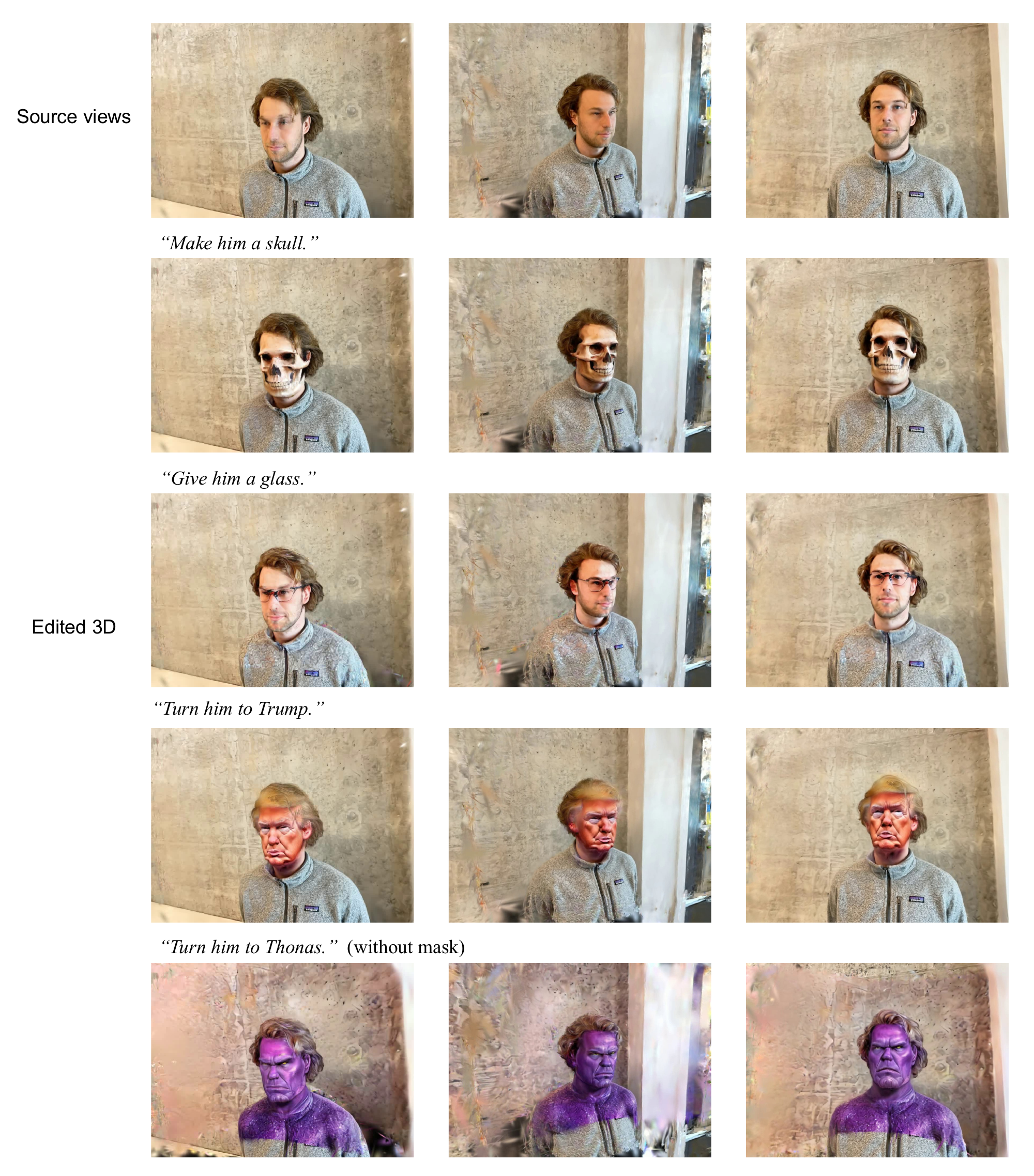}
        \vspace{-5pt}
        \caption{\textbf{Face.} } 
    \label{fig:suppl_vis_face}
\end{figure*}

\begin{figure*}[h]
    \centering
    \includegraphics[width=0.9\linewidth]{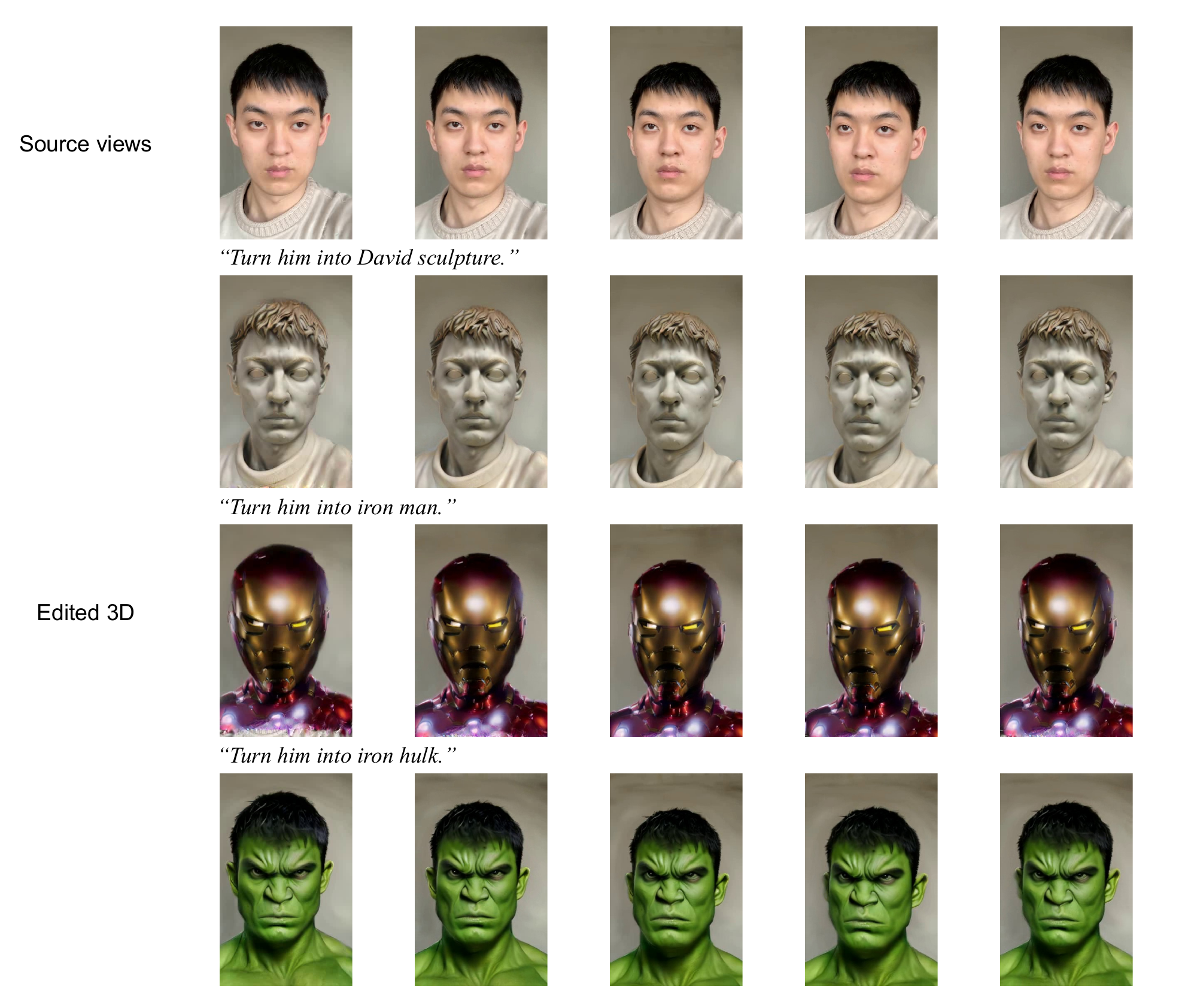}
        \vspace{-5pt}
        \caption{\textbf{Fangzhou.} } 
    \label{fig:suppl_vis_fangzhou}
\end{figure*}

\begin{figure*}[h]
    \centering
    \includegraphics[width=0.9\linewidth]{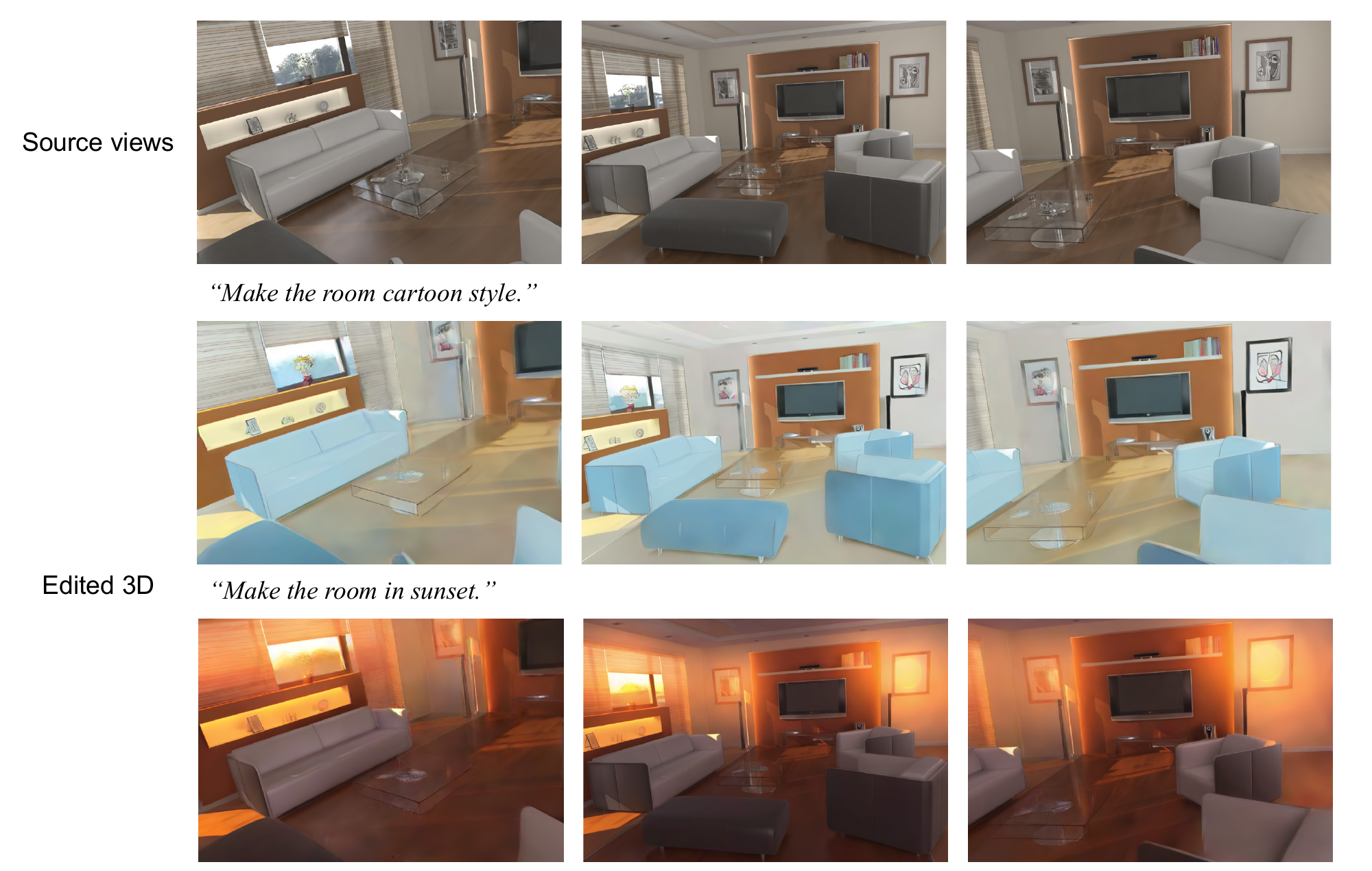}
        \vspace{-5pt}
        \caption{\textbf{Replica.} } 
    \label{fig:suppl_vis_replica}
\end{figure*}

\begin{figure*}[h]
    \centering
    \includegraphics[width=0.9\linewidth]{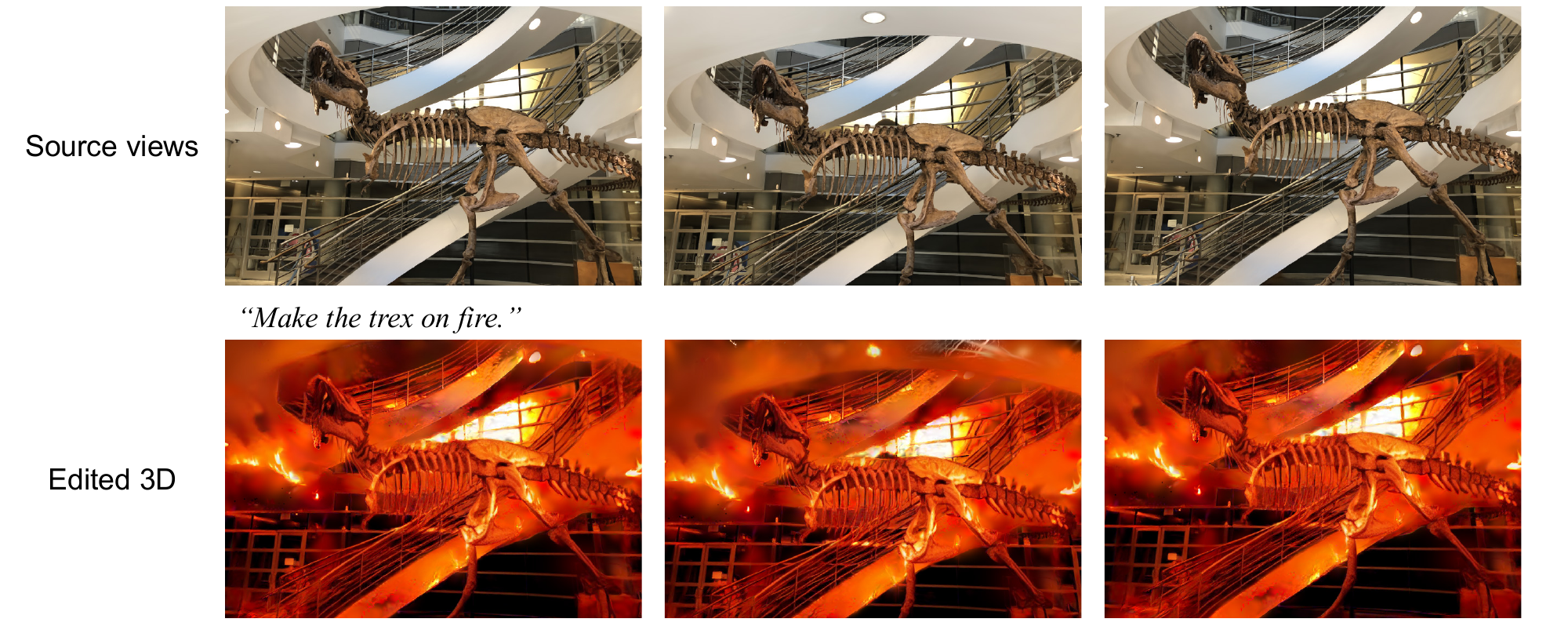}
        \vspace{-5pt}
        \caption{\textbf{Trex.} } 
        \vspace{-15pt}
    \label{fig:suppl_vis_trex}
\end{figure*}

\begin{figure*}[h]
    \centering
    \includegraphics[width=0.9\linewidth]{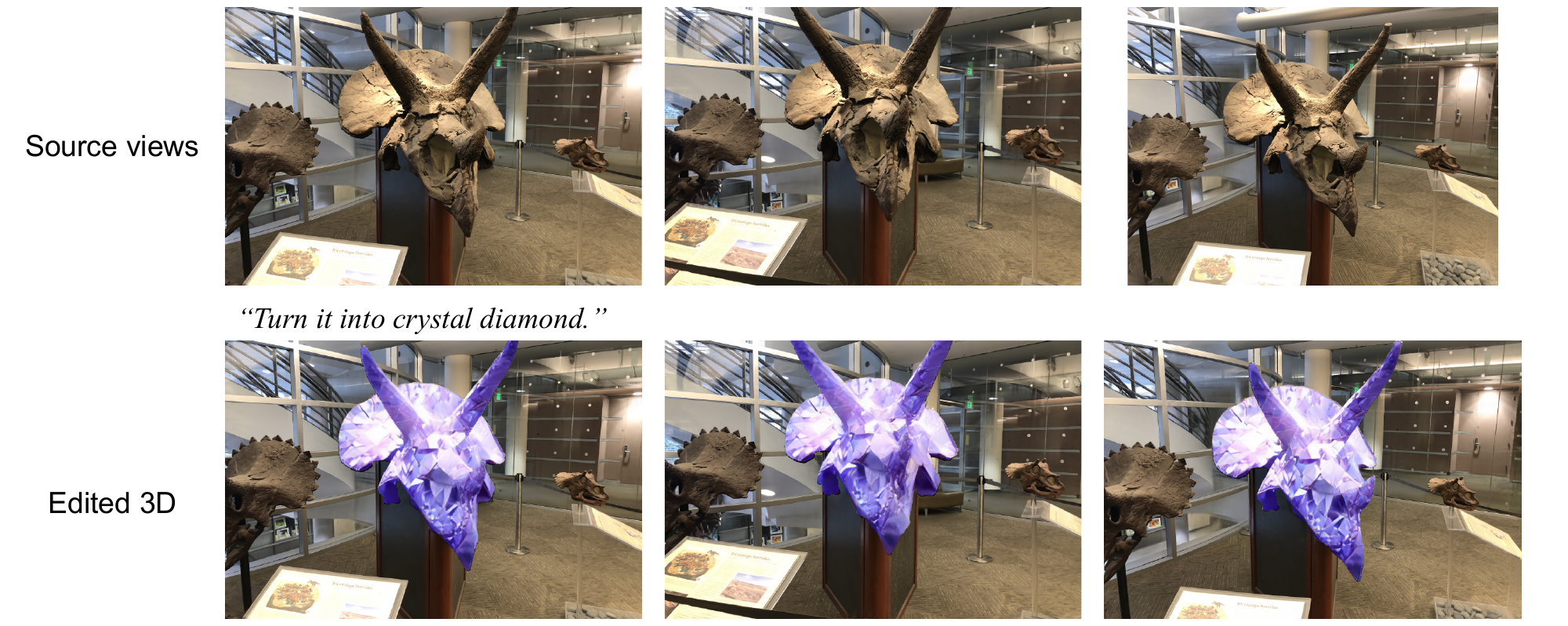}
        \vspace{-5pt}
        \caption{\textbf{Horn.} } 
        \vspace{-15pt}
    \label{fig:suppl_vis_horn}
\end{figure*}

\begin{figure*}[h]
    \centering
    \includegraphics[width=0.9\linewidth]{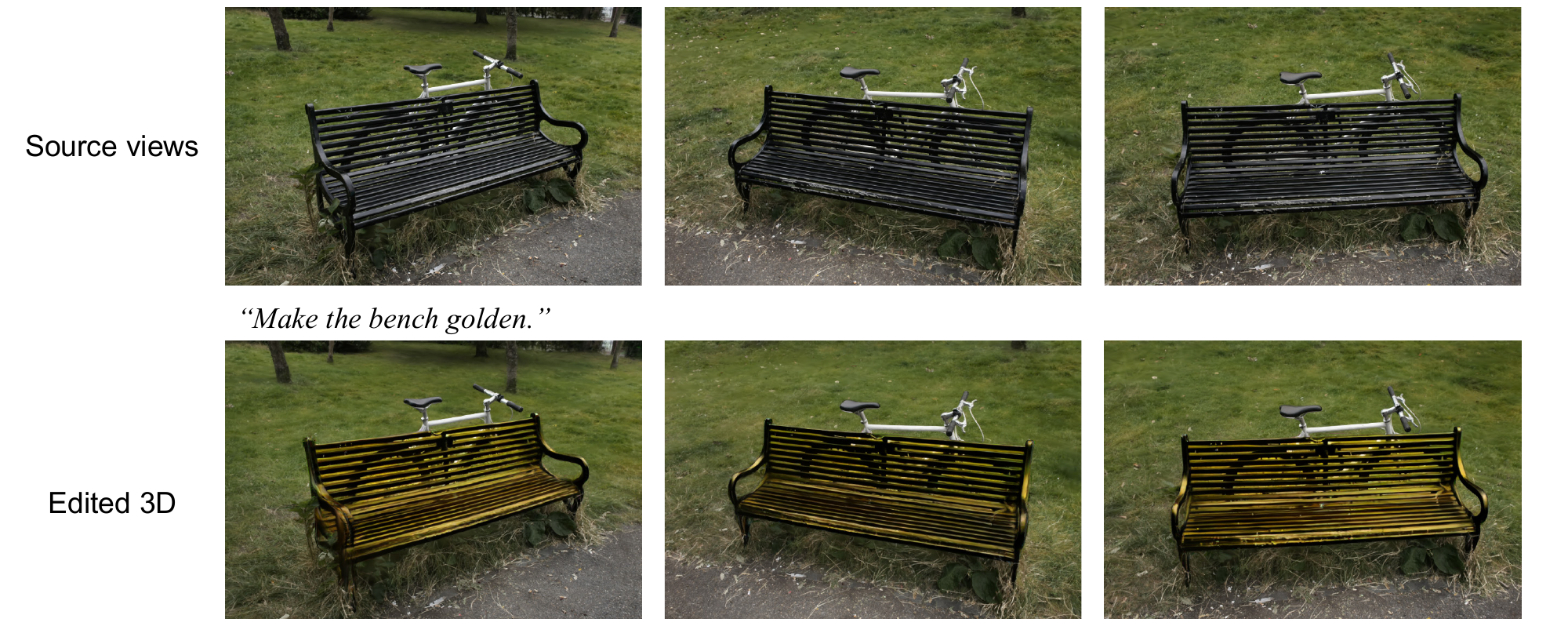}
        \vspace{-5pt}
        \caption{\textbf{Bicycle.} } 
    \label{fig:suppl_vis_bicycle}
\end{figure*}

\noindent \textbf{Visualization.}
We provide more visualization results to demonstrate the effectiveness of ViP3DE from Fig.~\ref{fig:suppl_vis_face} to Fig.~\ref{fig:suppl_vis_bicycle}. One can observe that even with a single forward pass, ViP3DE presents an impressive 3D consistency without blurs.

\clearpage

\clearpage

{
    \small
    \bibliographystyle{ieeenat_fullname}
    \bibliography{main}
}

\end{document}